\documentclass{clv3}

\usepackage{hyperref}
\usepackage{xcolor}
\definecolor{darkblue}{rgb}{0, 0, 0.5}
\hypersetup{colorlinks=true,citecolor=darkblue, linkcolor=darkblue, urlcolor=darkblue}

\definecolor{cite}{rgb}{0.6,0.6,1.0}

\definecolor{todo}{rgb}{1,0.5,0}

\usepackage{booktabs}
\usepackage{multirow}

\usepackage{floatrow}
\newfloatcommand{capbtabbox}{table}[][\FBwidth]

\issue{1}{1}{2016}

\dochead{}%

\runningtitle{Intertextual Model of Peer Review}

\runningauthor{Ilia Kuznetsov et al.}

\begin{document}

\title{Revise and Resubmit: An Intertextual Model of Text-based Collaboration in Peer Review}

\author{Ilia Kuznetsov}%
\affil{UKP Lab \\ Department of Computer Science \\ Technical University of Darmstadt https://www.ukp.tu-darmstadt.de/}

\author{Jan Buchmann}
\affil{UKP Lab}

\author{Max Eichler}
\affil{UKP Lab}

\author{Iryna Gurevych}
\affil{UKP Lab}
\maketitle

\begin{abstract}
Peer review is a key component of the publishing process in most fields of science. The increasing submission rates put a strain on reviewing quality and efficiency, motivating the development of applications to support the reviewing and editorial work. While existing NLP studies focus on the analysis of individual texts, editorial assistance often requires modeling interactions between pairs of texts -- yet general frameworks and datasets to support this scenario are missing. Relationships between texts are the core object of the intertextuality theory -- a family of approaches in literary studies not yet operationalized in NLP. Inspired by prior theoretical work, we propose the first intertextual model of text-based collaboration, which encompasses three major phenomena that make up a full iteration of the review-revise-and-resubmit cycle: pragmatic tagging, linking and long-document version alignment. While peer review is used across the fields of science and publication formats, existing datasets solely focus on conference-style review in computer science. Addressing this, we instantiate our proposed model in the first annotated multi-domain corpus in journal-style post-publication open peer review, and provide detailed insights into the practical aspects of intertextual annotation. Our resource is a major step towards multi-domain, fine-grained applications of NLP in editorial support for peer review, and our intertextual framework paves the path for general-purpose modeling of text-based collaboration. Our corpus and accompanying code are publicly available.\footnote{\url{https://github.com/UKPLab/f1000rd}
}
\end{abstract}

\section{Introduction}

Peer review is a key component of the publishing process in most fields of science: a work is evaluated by multiple independent referees -- peers -- who assess the methodological soundness and novelty of the manuscript and together with the editorial board decide whether the work corresponds to the quality standards of the field, or needs to be revised and resubmitted at a later point. As science accelerates and the number of submissions increases, many disciplines experience reviewing overload that exacerbates the existing weaknesses of peer review in terms of bias and reviewing efficiency. Past years have been marked by an increasing attention to the computational study of peer review with NLP applications ranging from review score prediction \cite{peerread} to argumentation analysis \cite{ampere}, and even first experiments in fully automatic generation of peer reviews \cite{reviewadvisor}.

We define \emph{text-based collaboration} as a process in which multiple participants asynchronously work on a text by providing textual feedback and modifying text contents. Peer reviewing is a prime example of text-based collaboration: similar to other fields of human activity, from lawmaking to business communication, paper\footnote{In this work we do not make distinctions between publication types; the terms "paper", "article", "manuscript" etc. are thereby used interchangeably.} authors receive textual feedback on their document, process it, modify the paper text accordingly and send it for another revision round. During this work, the participants need to address a range of questions. Is the feedback constructive, and what parts of the feedback text require most attention? What particular locations in the text does the feedback refer to? What has changed in the new version of the text, have the issues been addressed, and what parts of the text need further scrutiny? Answering these questions is hard, as it requires us to draw cross-document, \emph{intertextual relations} between multiple, potentially long texts. Despite the great progress in finding single documents and extracting information from them, cross-document analysis is only starting to get traction in NLP \cite{cdlm}. General frameworks and models to support text-based collaboration are yet to be established.

Treating text as an evolving entity created by the author and interpreted by the reader in the context of other texts is the core of the \emph{intertextuality theory} -- a family of approaches in literary and discourse studies not yet operationalized in NLP \cite{kristeva, broich, genette, steyer}. While the theoretical groundwork of the past decades provides a solid basis for such operationalization, existing body of work is mostly dedicated to the literary domain and lacks terminological and methodological unity \cite{quantint}. Inspired by the theoretical work, in this article we propose a  \emph{joint intertextual model of text-based collaboration} that incorporates three core phenomena covering one full document revision cycle: (1) \emph{Pragmatic tagging} classifies the statements in text according to their communicative purpose; (2) \emph{Linking} aims to discover fine-grained connections between a pair of texts; (3) \emph{Version alignment} aims to align two revisions of the same text. Creating this model requires us to revise the notion of text commonly accepted in NLP, and to adopt a new \emph{graph-based data model} that reflects document structure and encompasses both textual and non-textual elements crucial for human interpretation of texts. Our proposal is coupled with an implementation that allows extending the proposed data model to new document formats and domains.

Peer review is used by most fields of science, and both reviewing standards and publishing practices show significant variation across research communities. For example, the temporal restrictions of conference peer review reduce the number of potential revisions a manuscript might undergo; the continuity of research journals, on the other hand, allows the manuscript to be reviewed, revised and resubmitted multiple times before acceptance. Pre-publication review assumes that only accepted manuscripts get indexed and distributed; post-publication review happens after the publication, making accessible the papers that would otherwise be discarded. Finally, while most of the peer reviewing is closed and anonymized, some communities opt for open review, incl. the disclosure of the identities of authors and reviewers.

All these factors have substantial effects on the composition of the peer reviewing data, along with discipline-specific reviewing practices and quality criteria. However, existing NLP datasets of peer reviews \cite[etc.]{peerread, ampere, deepsentipeer} are exclusively based on conference-style, pre-publication, closed review in machine learning. To address this gap, we introduce the \emph{F1000Research Discourse corpus} (F1000RD) -- the first multi-domain corpus in journal-style, post-publication, open peer review based on the F1000Research platform.\footnote{\url{https://f1000research.com}} Based on this corpus, we instantiate our intertextual model in the peer reviewing domain and conduct annotation studies in pragmatic tagging, linking and version alignment for peer reviews and research papers, producing a novel \emph{multi-layered dataset} and providing \emph{key insights} into the annotation of intertextual phenomena. We finally apply our proposed framework to investigate the interaction between different types of intertextual relations. Our new resource is a major step towards multi-domain applications of NLP for peer reviewing assistance, and an exemplary source to guide the development of general models of text-based collaboration in NLP. 

In summary, this work contributes:

\begin{itemize}
	\item A theoretically inspired intertextual model of \emph{text-based collaboration};
	\item A new graph-based \emph{data model} based on an extended notion of text;
	\item Three novel \emph{tasks}: pragmatic tagging, linking, and version alignment;
	\item A richly annotated \emph{corpus} of multi-domain journal-style peer reviews, papers and paper revisions;
	\item Practical \emph{insights} and analysis of intertextual phenomena that accompany text-based collaboration during peer review.
\end{itemize}

The rest of the article is organised as follows: Section \ref{sec:bg} provides the necessary background in NLP for peer reviewing and text-based collaboration and introduces intertextuality theory and the dimensions of intertextuality that guide the development of our proposed framework. Section \ref{sec:framework} discusses the notion of text, introduces a novel graph-based data model well-suited for intertextual analysis, and formally specifies our proposed intertextual framework. This framework is instantiated in a corpus-based study of intertextuality in peer review in Section \ref{sec:corpus}, where we introduce the peer reviewing workflow of F1000Research, describe the F1000RD corpus, and provide detailed insights into our annotation studies and their results. Section \ref{sec:discussion} follows up with a more general discussion of future research directions. Section \ref{sec:conclusion} concludes the article with final remarks.

\section{Background}
\label{sec:bg}

\subsection{NLP and Peer Review}
\label{sec:bg:nlppeer}

Multiple strands of NLP research aim to improve the efficiency and fairness of peer review. A long-standing line of work in reviewer matching aims to generate reviewer-paper assignments based on the reviewers' expertise \cite{rpmatching1, rpmatching2}. Pioneering the use of NLP for peer review texts, \citet{peerread} introduce the tasks of review score and paper acceptance predictions, sparking a line of follow-up work \cite{deepsentipeer, dycke}. To compare reviewing practices between different communities, \citet{ampere} define a task of argumentation mining for peer reviews and devise a model which they then use to compare the composition of peer review reports at several major AI conferences. Recent works by \citet{reviewadvisor} and \citet{reviewrobot} pioneer the field of automatic peer review generation.

While work on review score and acceptance prediction based on the whole review or paper text is abundant, the applications of NLP to assist the process of reviewing itself are few: the argumentation mining approach of \citet{ampere} and the aspect and sentiment annotation by \citet{reviewadvisor} can be used to quickly locate relevant passages of the review report (e.g. questions or requests); the author response alignment approach suggested by \citet{ape} can assist reviewers, authors, and editors in disentangling discussion threads during rebuttal. We are not aware of prior work in NLP for peer reviews that models pragmatic role of peer review statements, links peer reviews to their papers, or compares paper versions -- although those operations form the very core of the reviewing process and could greatly benefit from automation. 

Peer review as a general procedure is used in most fields of science, but the specific practices, domain and topic distribution, reviewing standards, and publication formats can vary significantly across research communities and venues. Despite the methodological abundance, from the data perspective existing work in NLP for peer reviews focuses on a narrow set of research communities in AI that make their reviewing data available via the OpenReview\footnote{\url{https://openreview.net}} platform. We are not aware of any multi-domain datasets of peer reviews that represent a diverse selection of research communities.

Finally, although documents are discussed and revised in most areas of human activity, from lawmaking to education, the complex communication that surrounds the creation of texts often remains hidden and scattered across multiple communication channels. Peer review is an excellent source for the study of text-based collaboration, as it involves multiple parties reviewing and revising complex documents as part of a standardised workflow, with a digital editorial system keeping track of their communication and the corresponding text updates. To be useful for the study of text-based collaboration, this data should be made available under a clear, open license, and should include peer review texts as well as paper revisions. 

A brief analysis of the existing sources of reviewing and revision data in NLP reveals that none of them meet those requirements: as of April 2022 the ICLR content published via OpenReview is not associated with a license; while NeurIPS publishes peer reviews for accepted papers, their revision history is not available; and while arXiv provides pre-print revisions and is clearly licensed, it does not offer peer reviewing functionality. Table \ref{tab:corpus_prds} summarizes the prior sources of peer reviewing data and compares them to F1000Research, a multi-domain open reviewing platform that we introduce in Section \ref{sec:corpus}.

 \begin{table}[]
     \centering
     \begin{tabular}{r|ccc|c}
     & ICLR & NeurIPS & arXiv & F1000Research \\
     \midrule
     	  license & unclear & unclear & varied & CC-BY/CC0 \\
          reviews & yes & yes & no & yes \\
          revisions & yes & no & yes & yes \\
          domains & CS/AI & CS/AI & multi & multi \\
     \end{tabular}
     \caption{Sources of peer reviewing data in NLP to date, and F1000Research. arXiv catalog covers physics, mathematics, computer science, statistics, and others. F1000Research hosts publications from a wide range of domains, from meta-science to medical case studies, see Section \ref{sec:corpus:link}}
     \label{tab:corpus_prds}
 \end{table}

\subsection{NLP and Text-based Collaboration}

Our work focuses on the three core aspects of text-based collaboration: pragmatics, linking, and revision. Pragmatic tagging aims to assign communicative purpose to text statements and is closely related to the work in automatic discourse segmentation for scientific papers \cite{argzoning, lauscher, csabstruct}. Linking draws connections between a text and its commentary, and is related to work in citation analysis for scientific literature: citation contextualization \cite{chandrasekaran2020overview} aims to identify spans in the cited text that a citation refers to, citation purpose classification aims to determine why a given work is cited \cite{cohan2019structural}. Version alignment is related to the lines of research in analysis of Wikipedia \cite{edits2, edits1} and student essay revisions \cite{litman1, litman2}. 

For all three aspects, existing work tends to focus on narrow domains and builds upon domain-specific task formulations: discourse segmentation schemata for research papers are not applicable to other text types; citation analysis is facilitated by the fact that research papers use explicit inline citation style, so the particular citing sentence is explicitly marked -- which is clearly not the case for most document types; Wikipedia revisions are fine-grained and only cover few edits at a time, while in a general case a document might undergo substantial change in-between revisions. Domain-independent, general task models of pragmatic tagging, linking, and version alignment are yet to be established.

Moreover, treating those tasks in an isolated manner prevents us from modeling the interdependencies between them. Yet those interdependencies might exist: a passage criticizing the text is more likely to refer to a particular text location, which, in turn, is more likely to be modified in a subsequent revision, and the nature of the modification would depend on the nature of the commentary. To facilitate the cross-task investigation of intertextual relationships, a \emph{joint} framework that integrates different types of intertextual relations is necessary -- however, most related work comprises individual projects and datasets that solely focus on segmentation, linking or revision analysis. 

Establishing a joint framework for the study of intertextual relations in text-based collaboration would require a systematic analysis of intertextual relations that can hold. One such systematization is offered by the intertextuality theory -- a family of works in literary studies that investigates the relationships between texts and their role in text interpretation by the readers, and we briefly review it below.

\subsection{Intertextuality}
\label{sec:bg:it}

 \begin{table}[]
     \centering
     \begin{tabular}{r|c|c}
     type & explicit & implicit \\
     \midrule
     archi & document structure, templates & genre standards, pragmatics\\
	 meta & hyperlinks, citations, direct reference & allusion, plagiarism, indirect reference \\
	 para & edit history, manual diff & description of changes\\
     \end{tabular}
     \caption{Examples of intertextual relations by type (archi-, meta- and paratextuality) and overtness: explicit or implicit. Example of direct reference:``\textit{the discussion on page 5, line 9}''. Example of indirect reference:``\textit{the discussion of fine-tuning approaches <somewhere in the paper>}''.}
     \label{tab:itx}
 \end{table}

Any text is written and interpreted in the context of other texts. The term "intertextuality" has been coined by \citet{kristeva} and has since been refined, transformed and reinterpreted by subsequent work. While the existence of intertextual relationships is universally accepted, there exists little consensus on the scope and nature of those relationships, and a single unified theory of intertextuality, as well as a universal terminological apparatus, are yet to be established \cite{quantint}. Based on the prior theoretical work, we distill a set of \textbf{dimensions} that allow us to systematize intertextual phenomena and related practical tasks, and form the requirements for our proposed intertextual framework.

Intertextual relations can be categorized into (1) \textbf{types}.
A widely quoted typology by \citet{genette} outlines five core intertextuality types: \emph{intertextuality$^G$} (homonymous term redefined by Genette) is the literal presence of one text in another text, e.g. plagiarism; \emph{paratextuality} is the relationship between the text and its surrounding material, e.g. preface, title, or revision history; \emph{metatextuallity} holds between a text and a commentary to this text, e.g. a book and its critique; \emph{hypertextuality} is loosely defined as any relationship that unites two texts together and is not metatextuality; \emph{architextuality} is a relationship between a text and its abstract genre prototype, e.g. what makes text a poem, a news article, or a peer review report.

Intertextual relations vary in terms of (2) \textbf{granularity}, both on the source and on the target text side. \citet{steyer} summarizes this relation in terms of four \emph{referential patterns}: a part of a text referring to a part of another text (e.g. quotation), a part of a text referring to a whole other text (e.g. document-level citation), a whole text referring to a part of the other text (e.g. analysis of a particular scene in a novel) and a whole text referring to a whole other text (e.g. a preface to a book). We note that while the specific granularity of a text "part" is of secondary importance from a literary point of view, it matters in terms of both linguistic and computational modeling, and finer-grained distinction might be desirable for practical applications.

Intertextual relations vary in terms of their (3) \textbf{overtness}. On the source text side, the intertextual nature of a passage might be signaled \emph{explicitly} (e.g. by quotation formatting or citation) or \emph{implicitly} (e.g. by referring to another text without any overt markers \cite{broich}). The degree of explicitness of the marker might vary depending on the granularity or the target passage: a text might be not referred to at all (which would constitute allusion or plagiarism), referred to as a whole (e.g. most citations in engineering sciences), referenced with page or paragraph indication (e.g. most citations in humanities or references to legal and religious texts), up to the level of individual lines and statements (e.g. in the fine-grained discussion during peer review). We note that the type of the overt marker does not need to match the granularity of the target passage: a research paper might refer to a particular sentence in the source work, but only signal it by a document-level citation.

The three dimensions of intertextuality -- type, granularity, and overtness -- form the basis of our further discussion and cover a wide range of intertextual phenomena. Table \ref{tab:itx} provides further examples of intertextual relations across domains and use cases, focusing on the three intertextuality types relevant for our work: architextuality (pragmatic tagging), metatextuality (linking), and paratextualtiy (version alignment). It both demonstrates the scope of phenomena a general-purpose intertextual framework could cover, and the connections between seemingly unrelated aspects of intertextuality.

\subsection{Data Models}

A data model specifies how a certain object is represented and defines the way its elements can be related and accessed. %
Data models differ in terms of their expressivity: while a more expressive data model naturally retains more information, in research a less expressive model is often preferable as it poses less requirements on the underlying data, and can thereby represent more object types in a unified fashion. This motivates the use of \emph{linear text} as a de facto data model in NLP: a research paper, a news post and a Tweet can all be converted into a linear sequence of characters, which can be left as is and accessed by character offset (as in most modern datasets) or further segmented into tokens and sentences which then become part of the data model, as in most "classic" NLP datasets like Penn Treebank \cite{ptb}, Universal Dependencies corpora \cite{ud} and others.

Yet, a closer look at the phenomena covered by the intertextuality theory reveals that heavily filtered linear text might not be the optimal data model for studying cross-document discourse. Lacking a standard mechanism to represent document structure, linear model of text is not well suited for modeling phenomena on varying granularity levels -- yet humans readily use text structure when writing, reading and talking about texts. The de facto data model does not offer standartized mechanisms to represent (or at least, preserve) non-textual information, like illustrations, tables and references -- yet those often form an integral part of the document crucial for text interpretation and the surrounding communication. Finally, while it is possible to draw cross-document connections between arbitrary spans in plain text ad hoc, standartized mechanisms for representing cross-document relations are missing. All in all, while well-suited for representing grammatical and sentence-level phenomena, the current approach to text preprocessing at least invites a careful reconsideration.

First steps in this direction can be found in recent work. A closely related example, \citet{s2orc} introduce the S2ORC data model that allows unified encoding of document metadata, structure, non-textual and bibliographic information for scientific documents. This information is then used in a range of studies in document-level representation learning \cite{specter}, citation purpose classification \cite{cohan2019structural}, and reading assistance \cite{scholarphi}, demonstrating that non-linguistic elements and document structure are crucial for both devising better representations of text and assisting humans in text interpretation. The S2ORC data model is tailored to the idiosyncrasies of scientific publishing and publication formats. Our Intertextual Graph model introduced below is an attempt on a more general, inclusive model of text that offers a standartized way to represent document structure, encapsulate non-textual content, and capture cross-document relations -- all while being applicable to a wide range of texts.

\section{Proposed Framework}
\label{sec:framework}

We now introduce our proposed intertextual framework for modelling text-based collaboration, that we later instantiate in our study of peer reviewing discourse. As our prior discussion shows, modelling text-based collaboration poses a range of requirements our framework should fulfil. Text-based collaboration is not specific to a particular domain and readily crosses domain and genre boundaries, so our framework needs to be (R1) general, i.e. not tied to the particularities of domain-specific applications and data formats. As text-based collaboration involves multiple texts, our framework should be able to (R2) represent several texts simultaneously and draw intertextual relations between them. The framework should be suited for representing intertextual relations at (R3) different levels of granularity, and (R4) allow drawing relations between textual and non-textual content. Finally, the framework should enable (R5) joint representation of different intertextual phenomena to facilitate the study of dependencies between different types of intertextuality. Our further discussion of the proposed framework proceeds as follows: the data model defines the representation of input texts, which is then used to formulate task-specific models for pragmatic tagging (architextuality), linking (metatextuality) and version alignment (paratextuality).

\subsection{Intertextual Graph (ITG)}

\begin{figure}
    \centering
    \includegraphics[width=\linewidth]{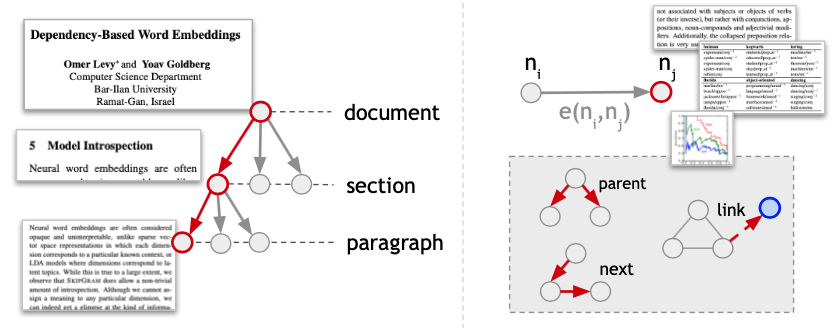}
    \caption{Basic Intertextual Graph. Left -- a full document ITG representing the logical structure of text on different levels via \texttt{parent} edges, with (Levy and Goldberg 2014) as example document. Right top -- nodes can encapsulate textual, as well as non-textual information, like tables and figures. Right bottom -- three core edge types in ITG. \texttt{link} edges can be divided in further subtypes depending on the relationship, can connect nodes of different modality (text and table) and granularity (sentence and section), and can cross document boundaries.}
    \label{fig:itg_basic}
\end{figure}

The core of our proposed framework is the Intertextual Graph graph data model (ITG, Figure \ref{fig:itg_basic}). Instead of treating text as a flat character or token sequence, ITG represents it as a set of nodes $N^G$ and edges $E^G$ comprising a graph $G$. Each node $n_i \in N^G$ corresponds to a logical element of the text. ITG supports both textual (e.g. paragraphs, sentences, section titles) and non-textual nodes (e.g. figures, tables and equations), allowing us to draw relationships between heterogeneous inputs. Nodes of the ITG are connected with typed directed edges $e \in E^G$. We denote an edge between nodes $n_i$ and $n_j$ as $e(n_i, n_j)$. We define three core edge types:

\begin{itemize}
    \item \texttt{next} edges connect ITG nodes in the linear reading order, similar to the mainstream NLP data models discussed above;
    \item \texttt{parent} edges mirror the logical structure of the document and the hierarchical relationship between sections, subsections, paragraphs, etc.;
    \item \texttt{link} edges represent additional, intertextual connections between graph nodes, ranging from explicit in-document references (e.g. to figure) to citations and implicit cross-document and version alignment links, as introduced below.
\end{itemize}

A node can have an arbitrary number of incoming and outgoing \texttt{link} edges, allowing the data model to represent many-to-many relationships; however, the \texttt{next} edges must form a connected list to represent the reading order, and the \texttt{parent} edges must form a tree to represent the logical structure of the text.

The proposed graph-based data model has multiple advantages over the traditional, linear representation of text widely used in NLP: it offers a standartized way to draw connections between multiple texts (R2); text hierarchy represented by the \texttt{parent} edges provides a natural scaffolding for modeling intertextual relations at different granularity levels (R3); a graph-based representation allows to encapsulate non-textual information while still making it available for intra- and intertextual reference (R4); the framework can represent different intertextuality types jointly, as we demonstrate later (R5). Our data model is generally applicable to any text (R1) as it does not rely on domain-specific metadata (e.g. abstracts, keywords) or linking and referencing behaviors (e.g. citations); the \texttt{next} edge mechanism retains the linear reading order of text and makes the documents represented within our model compatible with NLP approaches tailored to linear text processing, e.g. pre-trained encoders like BERT \cite{bert}, RoBERTa \cite{roberta}, Longformer \cite{longformer} etc.

Data models in NLP differ in terms of the units of analysis they enable; those units are used to devise representations of target phenomena: for example, syntactic parsing operates with tokens within one sentence, while text classification operates with whole documents. The hierarchical representation offered by the ITG allows flexibility in terms of the unit of analysis: in principle, an intra- or intertextual relationship can be drawn between a single token and a whole document (e.g. citations), or between a sentence and a table. While document is the largest unit of analysis in our data model, the definition of the smallest unit remains open and can be chosen on the application basis. Paragraph is the smallest unit of written text that does not require non-trivial preprocessing, and is an attractive first choice for the tasks where paragraph-level granularity is sufficient. However, most discourse analysis tasks, including pragmatic tagging and linking discussed below, require finer granularity, and in this work we use \emph{sentence as a minimal unit of analysis}. We point that this is a practical choice, and not a limitation of the proposed framework per se: the internal contents of a sentence can be represented as a sequence of word nodes connected via the \texttt{next} edges -- or as a character sequence referenced by offset, in accordance with the current mainstream approach to text representation.

\subsection{Architextuality and Pragmatics}

We now turn to pragmatic tagging as reflection of the text's architextual relationship to its genre. The original definition of the term by \citet{genette} is proposed in the context of literary studies and encompasses other genre-defining elements like style, form and declarative assignment (\textit{``A Poem''}); here we focus on the discourse structure as reflection of the genre. Text genres emerge as written communication standards, and to enable efficient communication, a text should adhere to the discourse structure imposed by its genre: a news article seeks to quickly convey new information, motivating the use of the pyramid structure where the most important new facts are placed in the beginning of the text; a research article, on the other hand, aims to substantiate a claim for new knowledge, and is expected to clearly delineate related work and state its own contribution; an argumentative essay takes a stance on a claim and provides evidence that supports or refutes this claim, etc. Being able to determine the pragmatic structure of a text is the first key step to its interpretation.

Unlike the other two relationships described below, architextuality holds not between a pair of texts, but between a text and its abstract ``prototype''. To reflect this, we introduce the task of pragmatic tagging, where an ITG node $n$ can be associated with a label from a pre-defined set $label(n) = l_i$, $\{l_1, l_2... l_j\} \in L$. Pragmatic structure of text can be signalled by the overt text structure, e.g. standardized section headings or text templates -- in which case the architextual relationship to the text genre is explicit; more frequently, however, it remains implicit and needs to be deduced by the reader. A pragmatic label can be associated with the units of varying granularity, from a text section (e.g. \textit{``Related work <...>''} $\rightarrow$ Background) to a particular sentence (e.g. \textit{``The paper is well written and the discussion is solid.''} $\rightarrow$ Strength).

Pragmatic tagging is a generalization of a wide range of discourse tagging tasks, incl. argumentative zoning \cite{argzoning} and subtasks of argumentation mining \cite{argmin1, argmin2, argmin3}, and is related to work in rhetorical structure analysis \cite{rst}. We note that our definition of pragmatic tagging does not cover the structural relationships between segments of the \emph{same} text (like argumentative structures in argumentation theory or rhetorical relations in rhetorical structure theory) -- while drawing such relations is easy within the proposed data model, this goes beyond the scope of our analysis. We instantiate pragmatic tagging in the peer reviewing domain by introducing a novel labeling schema for peer review analysis in Section \ref{sec:corpus:prag}.

\subsection{Metatextuality and Linking}

The next intertextuality type that our proposed framework incorporates is metatextuality, defined by Genette as a  \textit{``commentary... uniting a given text to another, of which it speaks without necessarily citing it''}. Despite its original use in the discussion of literary works, metatextuality lies at the very core of text-based collaboration, and spans beyond the literary domain: a book review, an related work survey, a social network commentary, or a forum thread post all participate in a metatextual relationship. Being able to draw such relationships between two texts is crucial for text interpretation -- however, as metatextual relationships are not always explicitly signalled, this might often present a challenge that can greatly benefit from analysis and automation.

To model metatextuality in our framework, we introduce the task of linking. Given two ITGs, the anchor graph $G^A$ and the target graph $G^T$, we use superscript notation to distinguish their nodes, e.g. $n_i^A \in N^A$. The goal of linking is then to identify the anchor nodes in $G^A$ and draw \texttt{link}-type edges between their corresponding target nodes in $G^T$, $e(n_i^A, n_j^T)$. Linking is a frequent phenomenon, and while some text genres enforce explicit linking behavior (e.g. citations in scientific literature), in most texts the linking is done implicitly (e.g. mentioning the facts from the target text). Contrary to Genette's definition, our definition of explicit linking subsumes the cases of direct text reuse via quotation. Links can vary greatly in terms of both source and target granularity: a sentence might link to a whole text or a particular statement in this text; and a paragraph of the anchor text might be dedicated to a single term mentioned in the target text. Links are frequently drawn between textual and non-textual content: for example, a sentence might refer to a table, and a social media post might comment on a video. While our work does not deal with multimodality, the encapsulation offered by the ITG data model enables such scenarios in the future.

The task of linking is a direct generalization of a wide spectrum of existing NLP tasks covering specific use cases and narrow domains: citation intent prediction \cite{cohan2019structural, teufel2006automatic}, citation contextualization \cite{chandrasekaran2020overview}, argument pair extraction \cite{ape}, review-paper mapping \cite{reviewrobot} and pairing comments on Wikipedia talk pages with the corresponding revision edits \cite{daxenberger-gurevych-2014-automatically} can all be cast as link-finding or link-labeling within our proposed framework. Linking relates to a broad category of evidence-finding NLP tasks like question answering with span prediction \cite{10.1162/tacl_a_00276, dasigi-etal-2021-dataset}, fact checking with evidence retrieval \cite{thorne-etal-2018-fever, wadden-etal-2020-fact}. Conceptually, linking is related to the work in cross-document structure theory (CST) \cite{radev-2000-common, radev-etal-2004-cst, Maziero2010-uj} which has been applied to short documents on sentence level in the newswire domain. While CST focuses on \emph{labeling} cross-document links and devising a typology of cross-document relations, in this work we focus on \emph{finding} links between documents, and instantiate implicit and explicit linking in the peer reviewing domain in Section \ref{sec:corpus:link}.

\subsection{Paratextuality and Version Alignment}

The final intertextuality type discussed here is paratextuality. \citet{genette} broadly defines paratextuality as a relationship between the core text of a literary work and the elements that surround it and influence the interpretation, incl. title, foreword, notes, illustrations, book covers, etc. We focus on a particular paratextual relationship highly relevant for modeling text-based collaboration -- the relationship between a text and its previous versions. An updated text is not interpreted anew, but in the context of its earlier verison; being able to align two versions of the same text is critical for efficient editorial work, as it would allow to quickly summarize the changes and highlight new material. Those time-consuming operations are mostly performed manually, as general-purpose models of text change are missing.

To address this, we introduce the task of version alignment. Given two Intertextual Graphs corresponding to the different versions of the \emph{same} text $G^t$ and $G^{t+\Delta}$, the goal is to produce an alignment, which we model as a set of intertextual edges $e(n_i^{t+\Delta}, n_k^t)$ between the two graphs\footnote{The direction of the version alignment is opposite to time (i.e. from $t+\Delta$ to $t$): a latter text refers to an earlier text, but not vice versa.}. Note that in our formulation the two versions of the document must not be consecutive, and the ability of the ITG to represent multiple documents allows us to simultaneously operate with multiple versions of the same document with an arbitrary number of revisions in-between. We further denote the revisions as \emph{short-scope} or \emph{long-scope} depending on the magnitude of changes between them. While not a strict definition, a typo correction constitues a short-scope edit, while a major rewrite constitues a long-scope edit.

In terms of overtness, the correspondence between two versions of the same text is rarely explicit: producing such alignment manually is time-consuming, and the logs that keep track of character-level edit operations are limited to few collaborative authoring platforms like Google Docs\footnote{\url{https://docs.google.com}} and Overleaf\footnote{\url{https://overleaf.com}}, and are too fine-grained to be used directly. The alignment between two text versions thereby remains implicit: while the general fact of text change is known, the exact delta needs to be discovered via ad-hoc application of generic utilities like \texttt{diff} or by requiring the text authors to provide a textual summary of changes and mark the updated parts of the text. Those might differ in terms of granularity from high-level notes (\textit{``We have updated the text to address reviewer's comments''} to in-detail change logs (\textit{``Fixed typos in Section 3; adjusted the first paragraph of the Introduction.''}); the selection of the granularity level depends on the application and the communicative scenario.

Version alignment is related to multiple strands of research within and outside NLP. Outside NLP, version analysis is explored in the software engineering domain \cite{codediff1, codediff2} -- which focuses on program code; related approaches based on simple text matching techniques exist in digital humanities, termed as \emph{collation} \cite{collation}. In NLP, Wikipedia edits and student essay writing have been the two prime targets for the study of document change. Both existing lines of research operate under narrow domain-specific assumptions about the nature of changes:  Wikipedia-based studies \cite{edits2, daxenberger-gurevych-2014-automatically} assume short-scope revisions characteristic to collaborative online encyclopaedia, and focus on edit classification, while essay analysis \cite{litman1, litman2} focuses on a narrow the case of student writing and medium-sized documents. Our task definition generalises from those previously unrelated strands of research and allows the study of long-scope long-document revisions, instantiated in the annotation study of research paper alignment in Section \ref{sec:corpus:ver}.

\subsection{Joint Modeling}

\begin{figure}
    \centering
    \includegraphics[width=0.6\linewidth]{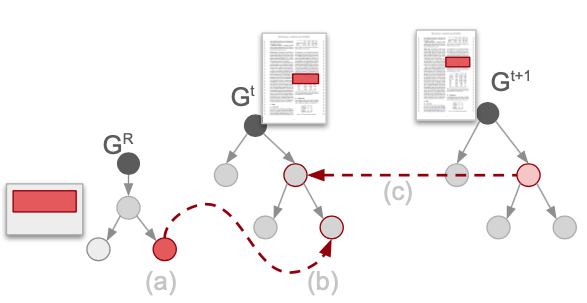}
    \caption{Joint modeling of multiple documents via ITG, simplified and with \texttt{next} edges omitted for clarity. A review $G^R$ node with certain pragmatics (a, pragmatic tagging) is connected (b, linking) to the main document $G^t$ which is later revised, producing a new version $G^{t+1}$ aligned to the original (c, version alignment). Following the links between documents enables new types of intertextual analysis -- yet the links are not always explicit (see Section \ref{sec:bg:it}), and might need to be inferred from text.}
    \label{fig:itg_joint}
\end{figure}

Apart from suggesting general, application-independent architectures for pragmatic tagging, linking and version alignment of arbitrary texts, our framework allows joint modeling of these phenomena (Figure \ref{fig:itg_joint}). Different types of intertextuality indeed interact: the communicative scenario that a text serves does not only prescribe its pragmatic structure, but also determines the standards of linking and the nature of updates a text might undergo. On a finer level, joint modeling of pragmatics, linking and version alignment allows us to pose a range of new research questions. Are metatextual statements with certain pragmatics more likely to be linked, and do statements with a large number of links tend to belong to a certain pragmatic category? Can explicit, readily available intertextual signal -- document headings, citations and detailed, character-level change logs -- be used as auxiliary signal for uncovering latent, implicit intertextual relationships? What parts of texts are more likely to be revised, and which factors contribute to this?  Our proposed framework facilitates joint analysis of intertextuality outside of narrow application-driven scenarios like using research paper structure to boost citation intent classification performance in \cite{cohan2019structural}. We demonstrate this capacity in our study of peer reviewing discourse which we now turn to.

\section{Corpus Study in Peer Review}
\label{sec:corpus}

The research publication is the main mode of communication in the modern scientific community. As more countries, organizations and individuals get involved in research, the number of publications grows, and while significant progress has been achieved in terms of finding research publications \cite{specter, covidir} and extracting information from them \cite{scierc, trialstreamer, pwc}, core technologies for prioritizing research results and ensuring the quality of research publications are lacking. The latter two tasks form the core of peer review -- a distributed manual procedure where a research publication is evaluated by multiple independent referees who assess its soundness, novelty, readability and potential impact.

As a result of peer reviewing, the manuscript is accepted and published, or rejected and discarded, or revised and resubmitted for another reviewing round. During this process, the authors, reviewers and volume editors work together to improve the manuscript so that it adheres to the scientific and publishing standards of the field. The communication happens over text, making peer review a prime example of text-based collaboration. The process often takes place in a centralized digital publishing platform which stores a full log of the interactions between the participants, incl. initial manuscript draft, peer reviews, amendment notes, meta-reviews, revisions and author responses; this makes peer review a unique, rich data source for the study of intertextual relations. 

Due to the anonymity of the process, however, this data often remains hidden. As discussed in Section \ref{sec:bg:nlppeer}, the few existing sources of peer reviewing data used in NLP are insufficient to support the intertextual study of peer reviewing due to the gaps in domain and data type coverage and the lack of clear licensing. To approach this, we introduce F1000Research as a new peer reviewing data source for NLP. While meeting all of our requirements, the F1000Research platform has other substantial differences to the reviewing and publishing platforms previously used in NLP research on peer reviews. We briefly outline the reviewing process of F1000Research and highlight those differences below.

\subsection{Data Source: F1000Research}

F1000Research is a multi-domain open access journal with fully open post-publication reviewing workflow. It caters to a wide range of research communities, from medicine to agriculture to R package development. Unlike regular, "closed" conferences and journals, F1000Research publishes the manuscripts directly upon submission, at which point they receive a DOI and become citeable. After this, the authors or the F1000Research staff invite reviewers, who provide review reports and \emph{recommendations} to \texttt{approve}, \texttt{approve-with-reservations} or \texttt{reject} the submission. 

Reviewers are presented with guidelines and domain-specific questionnaires, but reviews themselves are in a free-text format. Authors can write individual author responses and upload a new version which is reviewed by the same referees, producing a new round of reports. This "revision cycle" can repeat until the paper is approved by all referees. However, the official "acceptance decision" step common to traditional journals and conferences is not required here: a paper might be rejected by its reviewers and still be available and citeable. Crucially for our task, the reviewing process at F1000Research is fully transparent, reviewer and author identities are public, and reviews are freely accessible next to the manuscript under an explicit CC-BY or CC-0 license. All the articles and reviews at F1000Research are available as PDF and as easy-to-process JATS XML\footnote{\url{https://jats.nlm.nih.gov}}, which allows us to avoid the noise introduced by PDF-to-text conversion and makes fine-grained NLP processing possible.

All in all, F1000Research provides a unique source of fine-grained peer reviewing data  so far overlooked by the NLP comumnity. In this work we focus on papers, paper revisions and reviews from F1000Research, and leave a thorough exploration of author responses and revision notes to future work.

\subsection{Corpus Overview}
\label{sec:corpus:co}

The full F1000RD corpus published with this work was crawled on \mbox{April 22th, 2021} from the F1000Research platform using the official API \footnote{\url{https://f1000research.com/developers}}. Source JATS XML files were converted into the Intertextual Graph representation as described in Section \ref{sec:framework}. We have collected peer review texts, papers and revisions for each paper available at F1000Research at the time of the crawl. The resulting full dataset contains $5.4$k papers, of which $3.7$k have reviews and $1.6$k have more than one version (Table \ref{tab:paper_stats}). This makes our resource comparable to the widely known PeerRead dataset \cite{peerread} that contains approx. $3$k papers with reviews. Table \ref{tab:review_stats} provides basic statistics for the peer review part of the dataset; the number of reviews is sligntly lower but comparable to PeerRead (10k). We note the high proportion of accepted papers in both datasets, however, the accept-with-reservations mechanic is specific to F1000Research and allows us to collect more critical reviews that contain actionable feedback.

We have selected a sample from the full dataset for the in-detail investigations described in the following sections. To avoid domain bias, we would like our sample to include contributions from different disciplines and publication types; experiments in pragmatic tagging and linking require publications that have at least one peer review for the first version of the manuscript; experiments in version alignment additionally require the manuscript to have at least one revision. While the latter criteria are easily met via filtering, F1000Research does not enforce an explicit taxonomy of research domains. Instead, F1000Research operates with \textit{gateways} -- collections of publications that loosely belong to the same research community\footnote{\url{https://f1000research.com/gateways}}. To ensure a versatile multi-domain and multi-format sample, for this work we have selected publications from the following gateways. While it is possible for a publication to belong to multiple gateways, we have only selected publications assigned to a single gateway.

\begin{itemize}
    \item Science policy research (\texttt{scip}) publishes manuscripts related to the problems of meta-science, peer review, incentives in academia etc.
    \item ISCB Community Journal (\texttt{iscb}) is the outlet of the International Society for Computational Biology dedicated to bioinformatics and computational biology.
    \item RPackage (\texttt{rpkg}) publishes new R software package descriptions and documentations; those undergo community-based journal-style peer review and versioning.
    \item Disease outbreaks (\texttt{diso}) contains research articles in the public health domain; many recent publications are related to the COVID pandemic, vaccination programs and public response.
    \item Medical case reports (\texttt{case}) are a special publication type at F1000Research, but do not constitute a separate gateway; they often describe a single clinical case or patient, often focusing on rare conditions or new methodology and treatment.
\end{itemize}

Tables \ref{tab:paper_stats} and \ref{tab:review_stats} compare our sample to the full corpus. As evident from Table \ref{tab:review_stats}, the study sample contains a lower proportion of straight \texttt{approve} reviews, focusing on the reports that are more likely to discuss the submission in detail, link to the particular locations in the submission, and trigger a revision. Table \ref{tab:bydomain} compares the gateways' contributions to our study sample. As it shows, the sample contains similar amounts of text for each of the gateways; the divergence among average manuscript length reflects the publication type differences, e.g. medical case reports contain 1.4k words on average, while scientific policy articles span an average of 3.7k words.

\subsection{Preprocessing and Annotation Setup}

In addition to converting source documents from JATS XML into the ITGs, the reviews in the F1000RD study sample have been manually split into sentences; similar to \citet{thelwall} we clean up review texts by removing template reviewing questionnaires included by the F1000Research reviewing interface. Since manually segmenting research papers into sentences would be too labor-intensive, for papers we rely on the automatic parses produced by \texttt{scispacy}\footnote{\url{https://allenai.github.io/scispacy/}}.

Most annotations reported in this work were performed by two main annotators, both fluent non-native English speakers pursuing a Masters degree and comfortable with reading academic texts. We aimed to recruit annotators with diverse research backgrounds: one annotator had a background in Environmental Engineering, another one has previously studied Business administration and was pursuing a degree in Data and Discourse studies. When discussing the annotation studies, the further text refers to these two annotators, unless explicitly specified. Additional annotations were performed by the authors of this work, fluent non-native English speakers with extensive expertise in academic reading, as well as authoring and receiving peer reviewing feedback, with background in theoretical linguistics, computer science, as well as computational and molecular biology. To study the effect of domain expertise on linking (Section \ref{sec:corpus:link}), we additionally involved two medical experts who had both completed their studies and had a full practicing year by the time of the annotation study.

The differences between pragmatic tagging, linking and version alignment motivate the different approaches we took to annotate those phenomena. Pragmatic tagging was cast as a sentence labeling task, annotated by the two main annotators supplied with a guideline, and adjudicated by an expert. The annotation of implicit links was assisted by a suggestion module to reduce the number of potential linking candidates and lower the cognitive load on the main annotators who had to simultaneously handle two long documents during the linking process. The annotation of version alignment was performed automatically and later manually verified by the expert annotators. The following sections describe our annotation process in detail.

\begin{table}[h]
    \centering
    \begin{tabular}{l|ccc|cc}
    & papers & \#words & \#sentences & +reviews &  +revisions  \\
    \midrule
    full & 5.4k & 17.4M & - & 3.7k &1.6k \\
    sample & 172 & 496K & 24.2K & 172 & 122 \\
    \end{tabular}
    \caption{Paper first version statistics for F1000RD full corpus vs sample, as well as the number of papers that have more than zero reviews for the first version, and more than one version. Number of words here and further is the lower bound estimated via whitespace tokenization.}
    \label{tab:paper_stats}
\end{table}

\begin{table}[]
    \centering
    \begin{tabular}{cccccc|c}
                & case & diso & iscb & scip & rpkg & total \\
         papers & 45 & 37 & 31 & 31 & 28 & 172 \\
         \#words & 63K & 118K & 111K & 117K & 88K & 496K \\
    \end{tabular}
    \caption{Manuscript statistics in the F1000RD sample by domain.}
    \label{tab:bydomain}
\end{table}

\begin{table}[h]
    \centering
    \begin{tabular}{l|c|ccc|cc}
    & reviews & approve & approve-w-r & reject & \#words & \#sentences  \\
    \midrule
    full & 8053 & .55 & .38 & .07 & 2M & - \\
    sample & 224 & .36 & .53 & .11 & 59K & 4.9K \\
    \end{tabular}
    \caption{Review statistics for base F1000RD full corpus vs sample, with ratios for approve, approve-with-reservations and rejecting reviews.}
    \label{tab:review_stats}
\end{table}

\subsection{Pragmatic tagging}
\label{sec:corpus:prag}

\begin{figure}
    \centering
    \includegraphics[width=\linewidth]{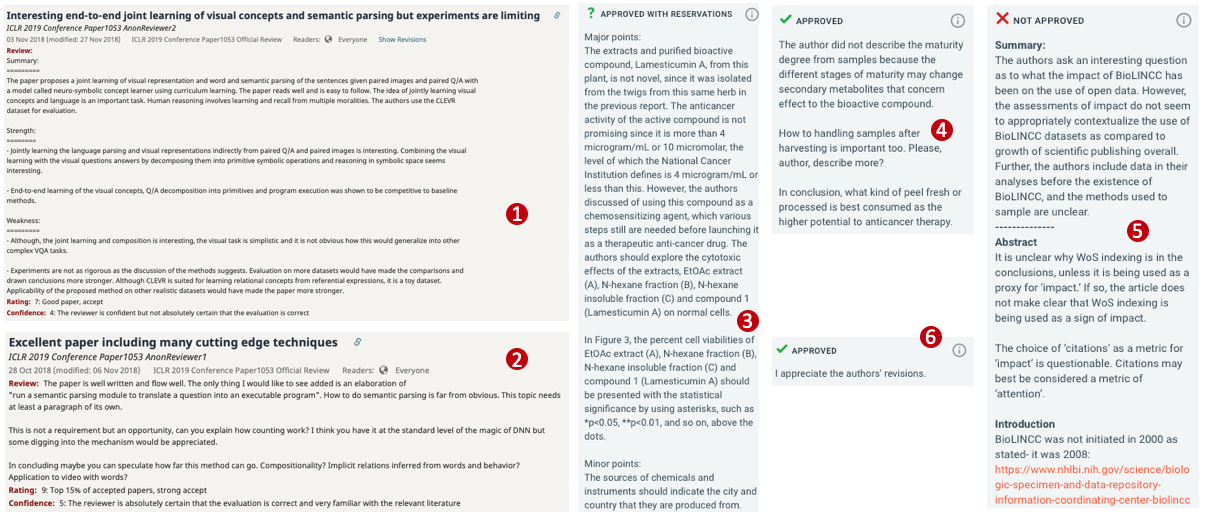}
    \caption{Diversity of peer reviewing styles and review report structures. Ex. \texttt{1} and \texttt{2} -- ICLR-2019 via OpenReview; Ex. \texttt{3-6} -- F1000Research.}
    \label{fig:prag_examples}
\end{figure}

While peer review is used in virtually any field of research, peer reviewing practices vary widely depending on the discipline, venue type, as well as individual reviewers' experience level, background and personal preferences. Figure \ref{fig:prag_examples} lists a range of reviews from ICLR-2019 (hosted by OpenReview) and F1000Research. As we can see, even within one venue reviews can differ dramatically in terms of length, level of detail and writing style: while the reports \texttt{1}, \texttt{3}, and \texttt{5} are structured, reports \texttt{2}, \texttt{4}, and \texttt{6} are free-form; moreover, among the structured reports, the reviewer in \texttt{1} groups their comments into summary, strengths and weaknesses; while reviewer in \texttt{3} organizes their notes by priority (major and minor points) and reviewer in \texttt{5} comments by article section. This illustrates the great variability of texts that serve as academic peer reviews.

Despite this variability, all peer reviewing reports pursue the same communicative purpose: to help the editor decide on the merit of the publication, to justify the reviewers' opinion and to provide the authors with useful feedback. Uncovering the latent discourse structure of free-form peer reviewing reports has several applications: it might help control the quality of reviewing reports by detecting outliers (e.g. reports that mention no strengths or do not provide a paper summary) before or after the report is submitted; it might help editors and authors navigate the peer review texts and summarize feedback; or it can be used to compare reviewing styles across disciplines and venues similar to the argumentation analysis by \citet{ampere}.

\subsubsection{Task and Annotation}

Motivated by this, we instantiate the task of pragmatic tagging $label(n) = l_i \in L$ in the peer reviewing domain with a sentence-level pragmatic tagging schema inspired by the related work in argumentation mining \cite{ampere} and sentiment analysis \cite{deepsentipeer, reviewadvisor} for peer reviews, as well as by the commonplace requirements from peer reviewing guidelines and templates. Our proposed six-class schema covers the major communicative goals of a peer reviewing report. \texttt{Recap} sentences summarize content of the paper, study, or resource without evaluating it; this includes both general summary statements and direct references to the paper. \texttt{Weakness} and \texttt{Strength} express an explicit negative or positive opinion about the study or the paper. \texttt{Todo} sentences contain recommendations and questions, something that explicitly requires reaction from the authors. \texttt{Structure} is used to label headers and other elements added by the reviewer to structure the text. Finally, an open class \texttt{Other} is used to label everything else: own reasoning, commentary on other publications, citations. Table \ref{tab:prag_label_ex} provides examples for each of the classes and compares them to AMPERE \cite{ampere}: while AMPERE focuses on surface-level argumentative structure, pragmatic analysis requires us to draw a distinction between Strengths and Weaknesses (both \textit{Evaluation} in AMPERE) and to separate the discussion of the background from the discussion of the manuscript under review (both \textit{Fact}). %

\begin{table}[]
    \centering
    \begin{tabular}{r|l|l}
    	label & example & AMPERE \\
    	\midrule
         Recap & \textit{The authors describe a case of <...>} & Fact \\
         Weakness & \textit{The figures are of low quality.} & Evaluation \\
         Strength & \textit{It is a well written software article.} & Evaluation \\
         Todo & \textit{Please specify whether <...>} & Request \\
         Structure & \textit{My major concerns:} & Other \\
         Other & \textit{As a non-surgeon, I can not <...>} & Other, Fact, Evaluation \\
    \end{tabular}
    \caption{Pragmatic tagging schema, examples and correspondence with the previously proposed AMPERE schema.}
    \label{tab:prag_label_ex}
\end{table}

To evaluate the robustness and coverage of our schema and produce the pragmatic tagging layer of the F1000RD corpus, we conducted an annotation study. After four hours of training, two annotators have labeled the F1000RD study sample according to the schema. While regular structured discussion meetings have been scheduled throughout the annotation process, the labeling itself was done independently by the two annotators, who reached a substantial \cite{agr} inter-annotator agreement of 0.77 Krippendorff's $\alpha$, demonstrating the high robustness of the proposed schema. Table \ref{tab:prag_iaa} outlines the inter-annotator agreement for pragmatic tagging by domain; as we can see, despite the domain differences, the schema remains robust across domains, suggesting that our proposed pragmatic categories of peer reviewing can be reliably detected and labeled universally. The initial annotations were adjudicated by an expert annotator (author of the schema), who resolved disagreements between annotators and in rare cases harmonized the annotation to fit the final guidelines taking into account the refinements made over the period of the annotation study. 

\begin{table}[]
    \begin{tabular}{c|ccccc}
         all & case & diso & iscb & rpkg & scip \\
         \toprule
         0.77 & 0.78 & 0.75 & 0.77 & 0.74 & 0.79
    \end{tabular}
    \caption{Inter-annotator agreement (Krippendorff's $\alpha$) for pragmatic tagging by domain.}
    \label{tab:prag_iaa}
\end{table}

\subsubsection{Analysis}

\begin{figure}
    \centering
    \includegraphics[width=0.6\linewidth]{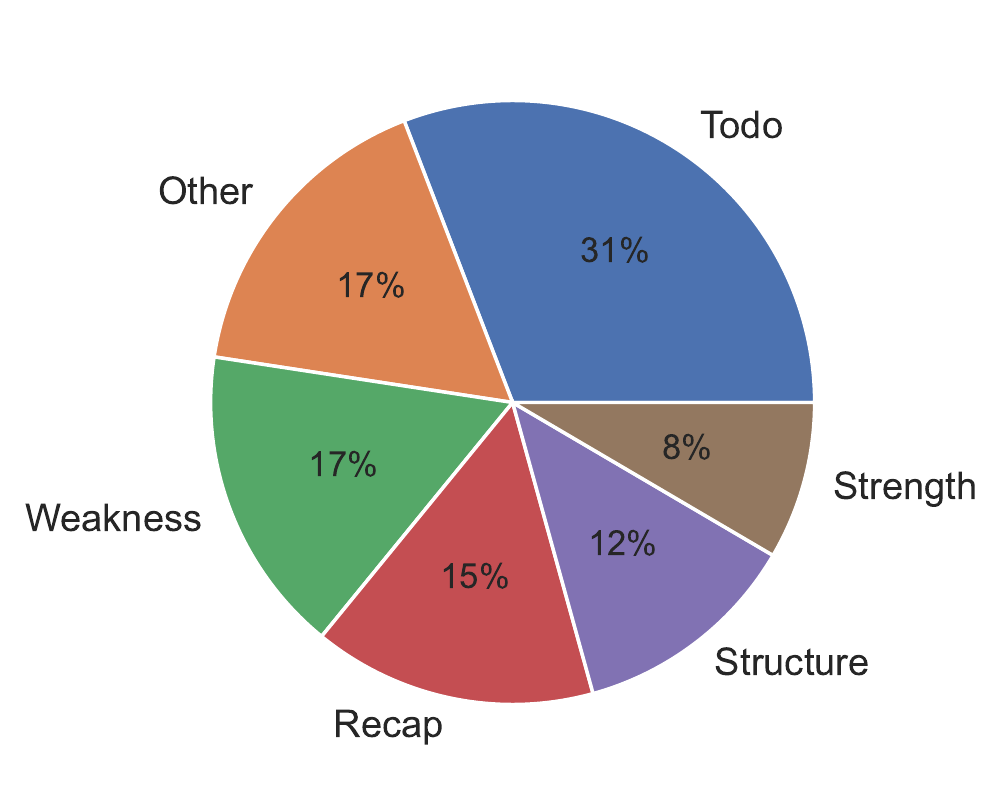}
    \caption{Distribution of pragmatic tags in F1000RD peer review annotations.}
    \label{fig:prag_dist}
\end{figure}

The resulting 4.9K labeled sentences make up the pragmatics layer of the F1000RD dataset. Figure \ref{fig:prag_dist} shows the distribution of pragmatic classes in the harmonized corpus. As we can see, the proposed annotation schema covers more than 82\% of the review full text, with only 17\% of the sentences falling into the catch-all \texttt{Other} category. Turning to the distribution of the core pragmatic classes, we note a high proportion of \texttt{Todo} sentences; while the related \texttt{Request} category in the AMPERE dataset makes up for less than 20\% of the data \cite[Table 3]{ampere}. One possible explanation for this discrepancy could be the difference between reviewing workflows: while AMPERE corpus builds upon conference-style reviewing data from ICLR with few revision rounds and quick turnaround, our data is based on journal-style, post-publication reviewing process of F1000Review, which allows the reviewers to make more suggestions for the next iteration of the manuscript. This finding highlights the importance of data diversity and calls for further comparative research in peer reviewing across domains, disciplines and publishing workflows.

Due to the robustness of the schema, the two annotators have agreed on the label for more than 80\% of the sentences in the corpus, creating a large catalog of clear examples of sentences for each of the pragmatic classes. Table \ref{tab:prag_examples_by_class} lists examples of clear cases by class; as it shows, the categories are natural enough to capture pragmatic similarities between sentences, while at the same time allowing for meaningful variation depending on the aspect in focus and the research domain. We note a difference in specificity and granularity of the sentences within the same pragmatic class, ranging from general statements (\textit{``The paper is hard to read''}) to surgical, low-level edit suggestions (\textit{``Implementation paragraph, line 6: pm-signature -> pmsignature.''}).

The common sources of disagreement between annotators highlight the limitations of the proposed schema and point at the directions for future improvement. We have observed a high number of disagreements between the \texttt{Recap} and \texttt{Other} categories due to the failure to distinguish between the manuscript itself, the accompanying materials, the underlying study, and background knowledge; especially pronounced in the medical and software engineering domains which make frequent references to the patient treatment and accompanying program code, respectively. Similarly, we have observed a large number of disagreements stemming from the interpretation of a sentence as neutral or evaluating: for example, \textit{``The authors have detailed the pattern seen in each case''} can be interpreted as a neutral statement describing the manuscript, or as a \texttt{Strength}. A important subset of the \texttt{Other} category not accounted for by our proposed schema is performative and meta-statements like \textit{``I recommend this paper to be indexed''} and \textit{``I don't have sufficient expertise to evaluate this manuscript''}. We note that such statements align well with the common elements of structured peer reviewing forms -- overall and confidence score -- highlighting the connections between explicit and implicit dimensions of the peer reviewing pragmatics.

\begin{table}[]
    \centering
    \begin{tabular}{l|l}
         Strength & XLA is a rare disease which gave this case report high value for being indexed.  \\
         & It is a well written case report and the discussion is precise. \\
& Each step is clearly explained and is accompanied by R code and output <...> \\
& I appreciated the author explaining the role that preprints could play <...> \\
& Recycling the water in the traps is a good idea in the short term because <...> \\
\midrule
Weakness & 1.5 year follow up is short for taste disorders. \\
& This doesn't seem efficient, especially with large single cell data sets such as <...> \\
& The use of English is somewhat idiosyncratic and requires minor review. \\
& The conclusions, while correct, are weak, and the results section is missing. \\
& The way they support their claim is problematic. \\
\midrule
Todo & How were Family vs Domain types handled from InterPro or Pfam? \\
& I recommend to the author to delete in the discussion the sentence <...> \\
& The following important reference is missing, please include: <...> \\
& The role of incentives in CHW performance should be discussed: <...> \\
    \end{tabular}
\caption{Clear-case examples of pragmatic classes from the corpus; some sentences are shortened and the \texttt{Other}, \texttt{Recap} and \texttt{Strcture} classes are omitted for the sake of presentation.}
\label{tab:prag_examples_by_class}
\end{table}

\subsection{Linking}
\label{sec:corpus:link}

Metatextuality is deeply embedded in academic discourse: each new study builds upon vast previous work, and academic texts are abundant with references to the corresponding publications; the number of incoming references accumulated over time serves as a proxy of the publications' influence, and the total reference count is a common measure of individual researchers' success. The main mechanism of intertextual referencing in academic writing is the citation; while the core function of connecting a text to a related previous text is common across research disciplines, the specific citation practices vary among communities, from document-level citations in textbooks to precise, page and paragraph-level inline references. Automatic analysis of citation behavior is a vast field of research in NLP \cite{cohan2019structural, teufel2006automatic, chandrasekaran2020overview}.

Like academic publications, peer reviews are also deeply connected to the manuscripts on an intertextual level. However, compared to full papers, peer reviews represent a much finer level of analysis as their main goal is to scrutinize a single research publication; moreover, since deep knowledge of the text is implied both from the author and from the reviewer side, most intertextual connections between the two texts remain implicit. Uncovering those connections bears great potential for automation and reviewing assistance, yet NLP datasets and methods to support this task are lacking.

\subsubsection{Task}

\begin{figure}
    \centering
    \includegraphics[width=0.7\linewidth]{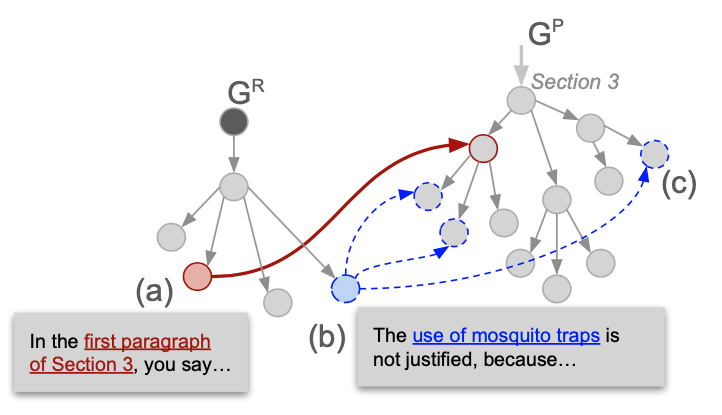}
    \caption{Linking between a review $G^R$ and a paper $G^P$. Only one paper section shown for simplicity, \texttt{next} edges omitted. While explicit linking (a) is facilitated by the presence of a clear anchor ("first paragraph") and clearly defined target in $G^R$, implicit linking faces two challenges at once, as it is unclear both whether the anchor sentence links to the paper at all (b), and if yes, what passages of the paper are linked to (c). Answering those questions requires simultaneous work with both documents; we employ a suggestion-based approach to make annotation feasible.}
    \label{fig:link_graphx}
\end{figure}

We instantiate the task of linking in the peer reviewing domain as follows: given the Intertextual Graphs of the peer review $G^R$ and the paper $G^P$, our goal is to discover the intertextual links $e(n_i^R, n_j^P)$ between the anchor nodes in the review $n_i^R \in G^R$ and the target nodes in the paper $n_j^P \in G^P$. We distinguish between explicit and implicit linking and model them as two separate subtasks. The anchor of an explicit link contains an overt marker pointing to a particular element of the paper text. It can also contain a clearly marked quotation from the paper, e.g. \textit{``In the Introduction, you state that <...>''}. The anchor of an implicit link refers to the paper without specifying a particular location, e.g. \textit{``The idea to set up mosquito traps is interesting''}, and is a substantially more challenging task. Similar to pragmatic tagging, we take sentence as a minimum unit for the anchor nodes $n_i^R$; the granularity of the target nodes $n_j^P$ is variable and depends on the subtask. Figure \ref{fig:link_graphx} illustrates the difference between the two kinds of linking, and Table \ref{tab:lnk_examples} shows examples from the F1000RD corpus.

\begin{table*}[t]
  \centering
    \begin{tabular}{p{5.4cm} c p{5.4cm}}
        \toprule
        Review Anchor & Link Type & Paper Target (Node Type) \\ \toprule
        The most important part of the article is in the \underline{discussion}.	& explicit & Discussion (section) \\ \midrule
        \multirow{3}{*}{\parbox{5cm}{\underline{Fig. 4} and \underline{Table 3}, interpretation is not helped by the lack of correspondence between names and code <...>}} & explicit  & Figure 4 (figure)  \\ \cmidrule{2-3}
        & explicit  & Table 3 (table) \\
        & & \\
        & & \\ \midrule
        It would be good to have a set of images from CellProfiler. & implicit  & Nuclei and infected cells were counted using CellProfiler. \mbox{(sentence)} \\ \midrule
        The authors intended to design a code requiring little R expertise.  & implicit  & We intentionally used simple syntax such that users with a beginner level of experience with R can adapt the code as needed.  \mbox{(sentence)} \\ \midrule
        \multirow{2}{*}{\parbox{5cm}{Details about the SVM learning algorithm must be included in the \underline{methods}.}} & explicit  & Methods (section) \\ \cmidrule{2-3} 
        & implicit  & SVM learning: Previously, paclitaxel-related response genes were identified, and their expression in breast cancer cell lines were analyzed by multiple factor analysis. (sentence) \\ 
    \end{tabular}
  \caption{Examples for explicit and implicit links. Explicit anchors are underlined.}
  \label{tab:lnk_examples}
\end{table*}

\subsubsection{Annotation: Explicit Links}

\begin{table*}[t]
  \centering
    \begin{tabular}{lrrrrrrrrrr|r}
    \toprule
         & lin  & pag & col & par & quo & sec & fig & tab & box & ref  & total \\ \toprule
    anchor  & 6  & 91   & 5 & 49  & 303   & 397   & 105   & 46   & 6   & 15     & 1023   \\
    links & -   & -   & - & 27   & 358   & 419   & 161   & 50   & 8   & 15 & 1038 \\
    \end{tabular}
  \caption{Distribution of explicit anchors and links in the F1000RD sample.}
  \label{tab:expl}
\end{table*}

The explicit linking layer in F1000RD was created in a two-step process. Based on an initial analysis of the corpus, we have compiled a comprehensive list of targets-of-interest for explicit linking, including line numbers, page numbers, columns, paragraphs (e.g. \textit{``second paragraph''}), quotes, sections, figures, tables and references to other work. Two authors of the study have manually annotated a random subset of 1100 peer review sentences with explicit link anchors and their types, reaching 0.78 Krippendorff's $\alpha$ on the general anchor/non-anchor distinction. Because of the good agreement, the rest of the data was manually annotated by one of the authors, who then also assigned target ITG nodes for each of the detected explicit anchors. Annotation was supported by a simple regular-expression-based parser reaching 0.77 and 0.64 F1 score for explicit link anchor and target identification on our annotated data, respectively. The regular-expression-based approach failed in cases such as unspecific quotes, imprecisions in the review texts (e.g. spelling errors) and other edge cases not handled by the rigid, hand-coded system.

Table \ref{tab:expl} shows the distribution of explicit anchor types and links in the resulting annotation layer: as we can see, explicit links are extensively used for referencing paper sections (sec), followed by literal quotes (quo) and figures (fig). Explicit links to lines (lin) and columns (col) are rare as the publication format in F1000Research is generally one-column and does not include line numbers. Page anchors (pag) are frequent -- yet publications are only split into pages during export, page numbers are not encoded in the source JATS XML files and thereby can't be linked to.

\subsubsection{Annotation: Implicit Links}

Compared to explicit links, a major challenge in implicit link annotation is the absence of both an overt intertextuality marker on the anchor side, and a clear attachment point on the target side. This requires the annotator to simultaneously perform anchor identification and linking on two potentially large, previously unseen texts. In a range of pilot studies, we have attempted to separate the task into anchor identification and linking, similar to explicit link annotation; however, our experiments have demonstrated low agreement on anchor identification solely based on peer review text in $G^R$. We have thereby opted for a joint annotation scenario with linking formulated as sentence pair classification task: given a pair of sentence nodes $n_i^R$ from the peer review and $n_j^P$ from the paper, the annotators needed to decide whether a link between the two sentences can be drawn.

While allowing the annotators simultaneous access to both texts, a pairwise classification setup inherently produces a large number of candidate pairs $|N^R| \times |N^P|$, most of which are irrelevant. To remedy this, similar to \citet{radev-etal-2004-cst} and \citet{mussmann-etal-2020-importance} we have implemented an annotation assistance system that given a review sentence $n_i^R$ presents the annotator with a suggestion set $S_i^P$ that consists of $m$ most similar paper sentences $n_j^P \in N^P$ which are subsequently annotated as linked or non-linked to the review sentence. To diversify the suggestion set, we construct it by aggregating the rankings from multiple similarity functions: for our annotation study we have used the BM-25 score \cite{INR-019} as well as cosine similarity between the review and paper sentence encoded using Sentence-BERT \cite{reimers-gurevych-2019-sentence} and Universal Sentence Encoder \cite{cer-etal-2018-universal}. When there were overlaps in the highest-ranked sentences from the different methods, the highest-ranked sentence not yet in the list of selected sentences was chosen iteratively for each method. 

Given the suggestions ($m = 5$), the annotators labeled the resulting sentence pairs according to the guidelines that specified the definition of a link and provided examples. Importantly, the annotators were asked to use paragraph level as the highest possible target granularity to avoid excessive linking of sentences that refer to the paper under review in general (e.g. \textit{``The overall writing style is good''}). Given the guidelines and the annotation support system, after four hours or training two annotators have labeled the F1000RD study sample. Even with annotation support, the task of implicit linking has proven substantially more challenging than pragmatic tagging, requiring twice as much time per review to produce annotations. The annotators have attained moderate \cite{agr} agreement of 0.53 Krippendorff's $\alpha$, and the resulting 21.289 labeled sentence pairs for 4.819 review sentences make up the implicit linking layer of the F1000RD dataset. Unlike for pragmatic tagging, given the moderate agreement we decided not to perform adjudication and enforce a single gold standard annotation, and instead release the two sets of labels produced by the annotators separately, similar to \citet{chandrasekaran2020overview} or \citet{fornaciari-etal-2021-beyond}.

\begin{table}[]
\centering
\begin{tabular}{lrrr}
\toprule
\multicolumn{1}{c}{ann} & \multicolumn{1}{c}{\#sent} & \multicolumn{1}{c}{\#item} & \multicolumn{1}{c}{$\alpha$}   \\ \toprule
\multicolumn{4}{c}{full}                                                                                                     \\
A, B                    & $4819$                          & $21289$                         & $0.53$                         \\ \hline
\multicolumn{4}{c}{re-test}                                                                                                  \\
A, B                    & \multirow{4}{*}{$809$}          & \multirow{4}{*}{$3630$}         & $0.53$                         \\
A$_{re}$, B$_{re}$                  &                                 &                                 & $0.56$                         \\
A, A$_{re}$                   &                                 &                                 & $0.51$                         \\
B, B$_{re}$                   &                                 &                                 & $0.58$                         \\ \hline
\multicolumn{4}{c}{expert}                                                                                                   \\
A, B                    & \multirow{2}{*}{$720$}          & \multirow{2}{*}{$3236$}         & $0.56^o$ / $0.59^m$ / $0.55^n$ \\
C, D                     &                                 &                                 & $0.48^o$ / $0.48^m$ / $0.47^n$ \\
\end{tabular}
\caption{Agreement statistics for implicit linking. The first column specifies the sets of annotations that are compared, where (A, B) are the annotations from the main study, (A$_{re}$, B$_{re}$) from the re-test and (C, D) from the experts. $o$, $m$, $n$: Agreement $o$verall, on $m$edical, on $n$on-medical.}
\label{tab:imp}
\end{table}

\subsubsection{Analysis} Our experiments demonstrate the large difference in complexity between the annotation of explicit and implicit links: while explicit linking can be performed semi-automatically, implicit linking requires extensive annotation assistance and presents a major conceptual challenge. Determining the source of complexity in annotation of implicit links is hard: related annotation studies by \citet{chandrasekaran2020overview}, \citet{10.1093/llc/fqs033}, \citet{reviewrobot} do not report inter-annotator agreement and don't investigate the factors contributing to the disagreements. We have identified three such possible factors and conducted additional experiments to investigate their impact.

\paragraph{Domain expertise} Our corpus includes scientific papers and their peer reviews from a wide range of domains, which might pose challenges both due to academic writing style and due to the domain knowledge required. While the high agreement on the pragmatic tagging might suggest that our annotators are not affected by the domain, linking might require more intimate knowledge of the subject compared to pragmatic tagging. To investigate the impact of domain expertise on annotation performance, we recruited two additional expert annotators with strong medical background and conducted the implicit linking annotation study on a subset of F1000RD peer reviews and papers in the medical domain (\texttt{case} gateway), using the same protocol and guidelines as our main study. We then compare the agreement of the experts in their domain of expertise and in other domains: if the disagreements are indeed due to the lack of domain knowledge, we expect to see a higher agreement between the expert annotators in their domain of expertise. As Table \ref{tab:imp} shows, this is not the case: moreover, we observe \emph{lower} agreement levels between the experts than between our main annotators who received additional training, annotated more data, and participated in the conceptualization of the task. This suggests that broad domain knowledge plays a secondary role in annotator agreement for the implicit linking task. 

\paragraph{Subjectivity} Another potential source of low agreement is task subjectivity. Since implicit linking involves joint decision-making on anchor identification and target identification, it is vulnerable to disagreements on both whether the review sentence should be linked at all, and what paper sentence it should be linked to. To study the effect of subjectivity, we have conducted an additional experiment where after a substantial period of time (2 months) the main annotators have re-labeled a subset of documents from the F1000RD corpus. Such test-retest scenario allows us to measure not only the agreement levels between the two annotators, but also the self-agreement level. A moderate inter-annotator agreement with high self-agreement would signal that the task is conceptually easy, but highly subjective, since the annotators would disagree with each other, but conform to their own earlier decisions. As our results in Table \ref{tab:imp} show, this is not the case: while the inter-annotator agreement is slightly improved in the retest, the self-agreement values stay in the same moderate-agreement range as in the main annotation study. This suggests that the disagreements on implicit linking do not stem from the task subjectivity per se.

\paragraph{Task conceptualization} The lack of domain expertise effect and the moderate self-agreement point at task definition as a potential target for further scrutiny. Intended as exploratory, our study deliberately left room for interpretation of the linking task. If high inter-annotator consistency is the goal, a stricter boundary between links and non-links would perhaps lead to higher agreement -- and we see theoretical works in intertextuality theory as a promising source of inspiration for such delineation. We discuss other promising alternatives to our task definition and annotation procedure in Section \ref{sec:discussion}.

\subsection{Version Alignment}
\label{sec:corpus:ver}

\begin{figure}[h]
    \centering
    \includegraphics[width=\linewidth]{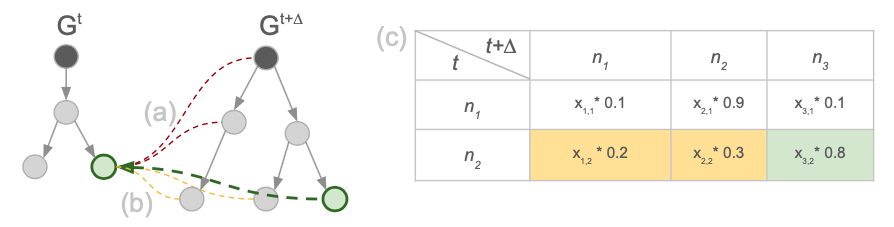}
    \caption{Version alignment, only potential edges for a single node shown for clarity. The ILP formulation penalizes aligning nodes that belong to different node types (a) and uses node similarities (b) to construct a global alignment that maximizes total similarity (c) under constraints.}
    \label{fig:ver_graph}
\end{figure}

Change is a fundamental property of texts: while any text undergoes changes as it is created, digital publishing makes it possible to amend texts even after publication, and this capability is widely used, from news to blog posts to collaborative encyclopediae. Academic publishing and text-based collaboration that surrounds peer review provide an excellent opportunity to study document change over time -- yet, while some tools like Google Docs and Overleaf make it possible to trace the document changes on character level, in most cases only major revisions of texts are exchanged; while some publishers require the authors to describe the changes, those descriptions are rarely enforced, not standardized, and not guaranteed to be complete. This makes it hard to verify whether the reviewers' expert feedback has been addressed, and to find out which parts of the manuscript are new and require attention; the performance of the ad-hoc solutions like \texttt{diff-match-patch}\footnote{\url{https://github.com/google/diff-match-patch}} on manuscript version alignment in the academic domain has not been systematically assessed.

\subsubsection{Task and Annotation} Motivated by this, we have conducted a study in automatic revision alignment of scientific manuscripts in the F1000RD sample. For simplicity, we cast the task as one-to-one ITG node alignment and only consider paragraph- and section-level alignment. Under those considerations, given two manuscript versions $G^{t+\Delta}$ and $G^t$ we aim to create a new set of edges $e(n^{t+\Delta}_i, n_j^t) \in E$ that signify the correspondences between revisions. Inspired by the work in graph-based annotation projection \cite{furstenauilp}, we formulate our approach as a constrained optimization via integer linear programming (ILP): in particular, given the set of nodes $(n_1^{t+\Delta}, n_2^{t+\Delta}, ..., n_i^{t+\Delta}) \in G^{t+\Delta}$ and $(n_1^t, n_2^t... n_j^t) \in G^t$, we define a binary variable $x_{i,j}$ that takes the value of $1$ if $n_i^{t+\Delta}$ is aligned to $n_j^t$, i.e. if we draw an edge $e(n^{t+\Delta}_i, n_j^t)$, and $0$ otherwise. We then seek to maximize the objective $\sum_i \sum_j x_{i,j} * score(n_i^{t+\Delta}, n_j^t)$ under the one-to-one alignment constraints $\forall j \sum_i x_{i,j} \leq 1$ and $\forall i \sum_j x_{i,j} \leq 1$. As part of the scoring function, we use Levenshtein ratio and word overlap to compute similarity between ITG nodes $sim$; in addition, we penalize the alignment of nodes that have different node types (e.g. paragraph and section title): $score(n_i^{t+\Delta}, n_j^t) = 0$ if $type(n_i^{t+\Delta}) \neq type(n_j^t)$, else $sim(n_i^{t+\Delta}, n_j^t)$. The result of the graph alignment is a set of cross-document edges connecting the nodes from $G^{t+\Delta}$ to $G^t$. Figure \ref{fig:ver_graph} illustrates our version alignment approach. The nodes in $G^{t+\Delta}$ with no outgoing alignment edges are considered added; the nodes in $G^t$ with no incoming edges from the future version of the document, are considered deleted.

The alignments produced this way\footnote{We used the PuLP library \url{https://coin-or.github.io/pulp/} as our ILP solver} were evaluated by three expert annotators, who were presented with aligned manuscript revisions and asked to judge the correctness of the node pairings. While it would be possible to include node splitting and node merging into our objective by modifying the ILP constraints, this remained beyond the scope of our illustratory study.

\begin{table*}[t]
  \centering
    \begin{tabular}{l|cc|ccccc}
    \toprule
         \#versions & full & sample & diso & iscb & rpkg & case & scip \\
         1 & 3743 & 50 & 10 & 18 & 14 & - & 8 \\
2 & 1353 & 105 & 22 & 9 & 12 & 42 & 20 \\
3 & 243 & 17 & 5 & 4 & 2 & 3 & 3 \\
4+ & 47 & - & - & - & - & - & - \\
         
    \end{tabular}
  \caption{Number of versions per paper in the full F1000RD corpus and the sample, by domain (Section \ref{sec:corpus:co}).}
  \label{tab:ver_detail}
\end{table*}

\subsubsection{Analysis} As the earlier Table \ref{tab:paper_stats} demonstrates, we oversample documents with revisions. A more detailed view in Table \ref{tab:ver_detail} shows that a significant number of documents in F1000RD undergo at least one revision before acceptance. We see that the number of revisions varies by the gateway, reflecting potential differences in publishing and reviewing practices across research communities.

\begin{figure}
    \centering
    \includegraphics[width=0.7\linewidth]{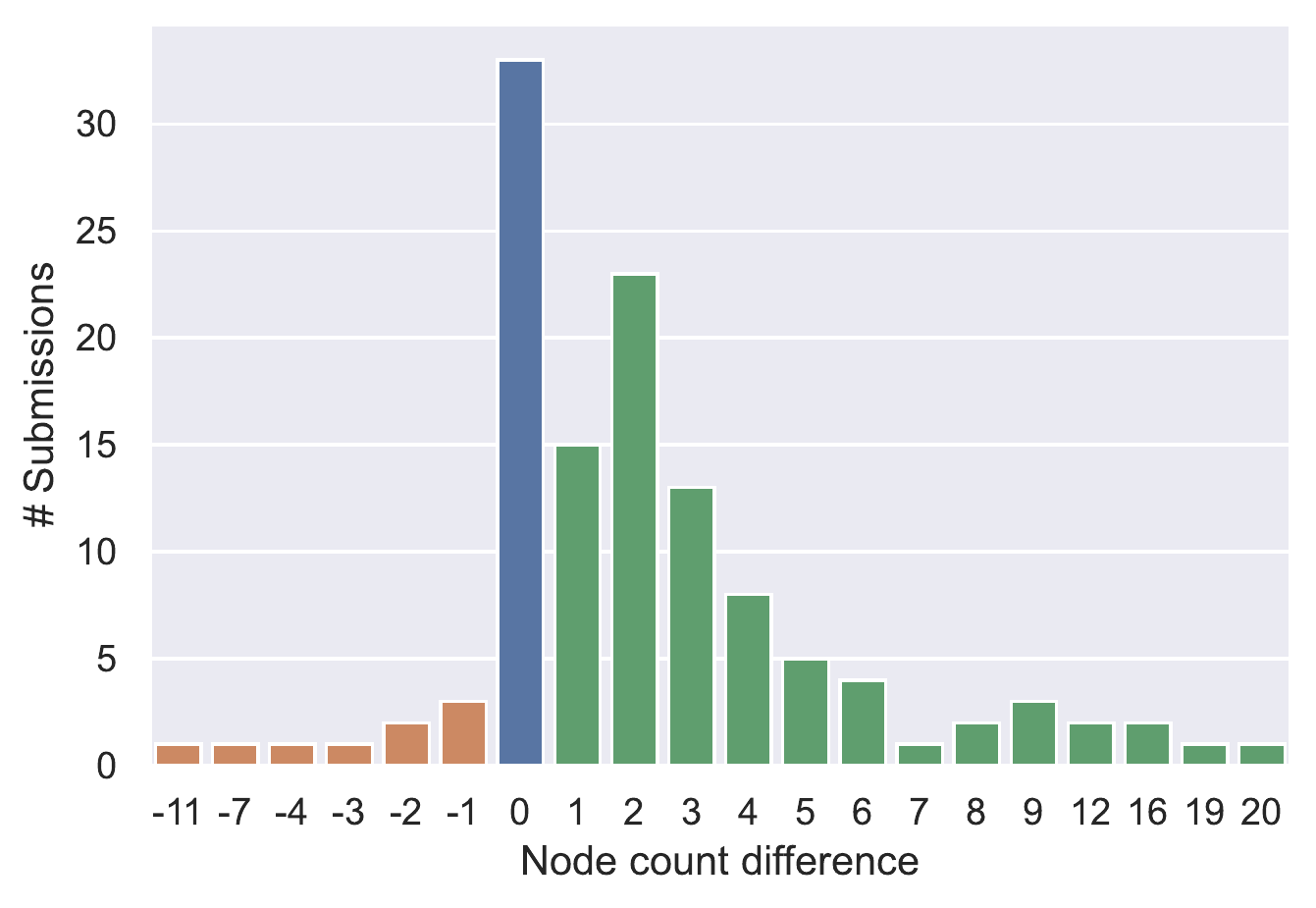}
    \caption{Difference in the number of nodes between submission and first revision in the F1000RD sample, total number of papers. Negative values mean that the revised version is shorter than the original submission.}
    \label{fig:ver_node_delta}
\end{figure}

For a more in-depth analysis, we focus on the differences between original submissions and their first revisions, as they are most numerous in our data. As Figure \ref{fig:ver_node_delta} demonstrates, in most cases the second revision of the manuscript contains more nodes, signifying incremental growth of the text in response to reviewer feedback. We note that the lack of change in the total number of nodes does not mean the lack of edits -- those edits simply do not affect the document structure. These results reflect another property of the F1000Research publishing workflow: the lack of a formal page limit allows the authors to add information without the need to fit and restructure the publication, which is often not the case with other publishers. This observation highlights the importance of taking the publishing workflow into account when working with the revision data.

\begin{table*}
\centering
\begin{tabular}{lccc}
\toprule
{} &  precision &  precision w/o exact &  perfect alignment \\
\midrule
Levenshtein distance (norm) &      0.982 &                0.966 &              0.713 \\
Word overlap      &      0.985 &                0.973 &              0.746 \\
\end{tabular}
\caption{Version alignment precision for submission and first revision in the F1000RD study sample, precision without exact matching, and the proportion of documents with perfect alignment. Only paragraph and section title-level nodes are considered.}
\label{tab:ver_accuracy}
\end{table*}

Finally, we turn to the quality of our approach to automatic version alignment. Table \ref{tab:ver_accuracy} shows the performance of our proposed alignment method when validated by the three expert annotators. As we can see, our simple ILP-based aligner reaches good alignment precision independent of the similarity metric; at the same time we note the lower number of documents where \emph{all} paragraph and title nodes have been correctly aligned, indicating room for improvement. While paragraph-level alignment is sufficient for our joint modeling study, more fine-grained analysis might be desirable for other tasks -- we discuss this and other further directions for the study of version alignment in Section \ref{sec:discussion}.

\subsection{Joint Modeling}

Together, pragmatic tagging, linking and version alignment allow us to cover one full reviewing and revision cycle of a common peer reviewing process. Each of the analysis types allows to answer practical questions relevant to text based collaboration: once automated, pragmatic tagging can help to quickly locate relevant feedback and analyse the composition of the reviewing reports; linking allows navigation between peer reviews and their papers, and version alignment allows easy comparison of document revisions. The joint representation of the three phenomena by ITG allows us to explore additional questions that provide deeper insights into text-based collaboration during peer review.

\paragraph{Data preparation} For each paper in the F1000RD sample, we construct an ITG for the paper itself, its reviews and its first revision. We aggregate pragmatic tagging annotations for the review, explicit and implicit linking edges between the review and the paper, as well as version alignment edges. We limit implicit links to the cases where both annotators agree. Note that while pragmatic tagging and implicit linking are performed on sentence level, version alignment happens on paragraph and section level, and the granularity of explicit links might vary. The ability of the ITG to handle different granularities allows us to integrate these annotations: while pragmatic tagging remains on the sentence level, linking annotations are propagated to paragraph granularity to make them interoperable with version alignments, see Figure \ref{fig:itg_joint}.

\begin{figure}
    \centering
    \includegraphics[width=0.7\linewidth]{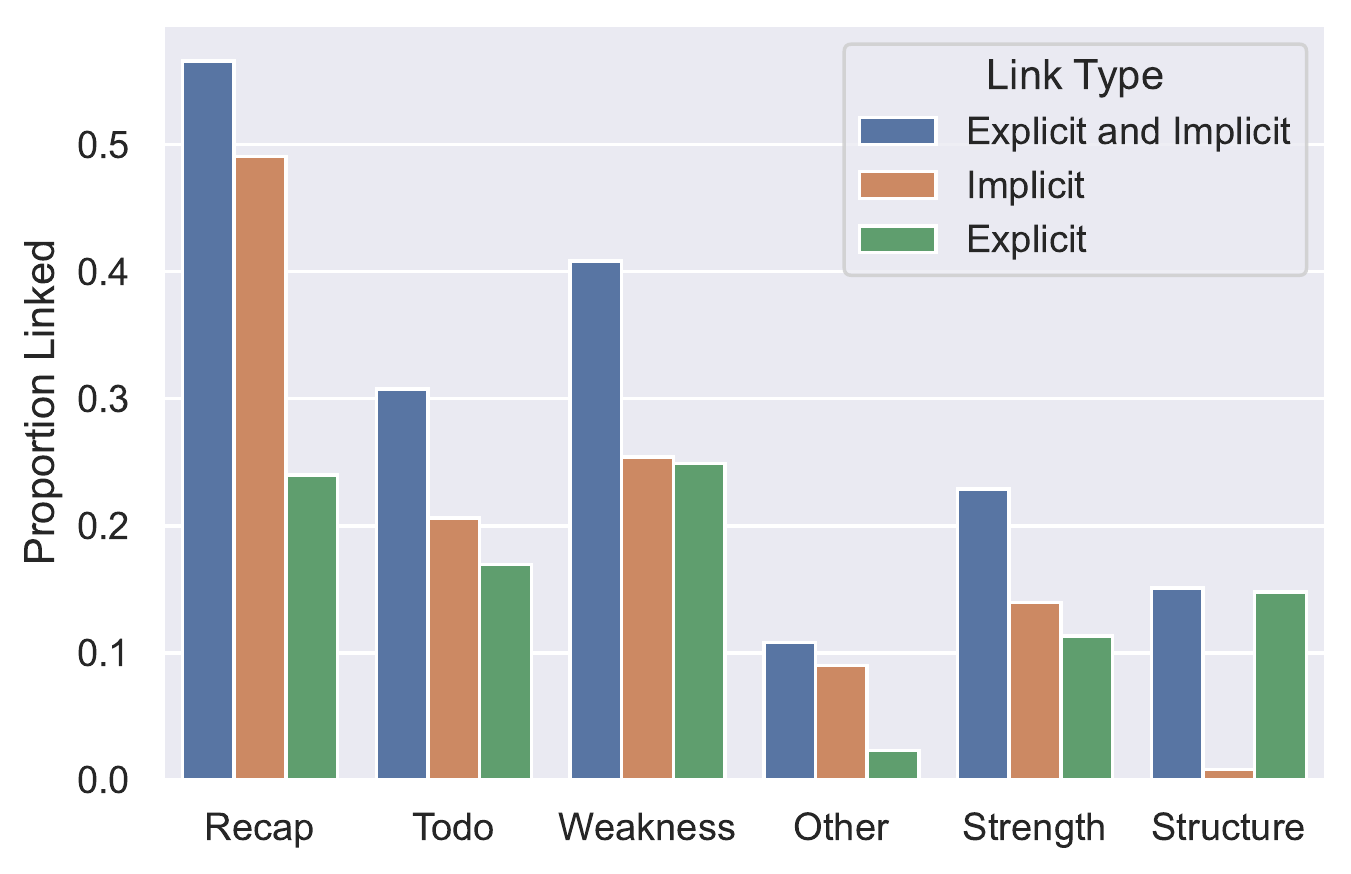}
    \caption{Pragmatics of peer reviews and the linking behavior.}
    \label{fig:joint_prag_link}
\end{figure}

\paragraph{Why do reviewers link} Reviewers use implicit and explicit links to refer to the paper they discuss -- yet one might expect that the use of links is not uniform. We use our data to investigate the interaction of review pragmatics and linking, and find out why reviewers link. As Figure \ref{fig:joint_prag_link} shows, there is indeed a clear dependency between linking behavior and pragmatics of peer review reports: for example, \texttt{Weaknesses} get linked almost twice as often as \texttt{Strengths}, potentially reflecting the fact that while the praise tends to address the work in general, criticism is likely to point at particular locations in the manuscript that need improvement. \texttt{Recap} is rarely linked to the work explicitly, but often rephrases and summarizes the paper content, producing an implicit link. \texttt{Todo} sentences are only linked to the manuscript in 1/3 of cases, pointing at a potential target for improvement in the reviewing guidelines -- yet, we note that a \texttt{Todo} might be related to a prior Weakness statement that contains a link, reflecting a limitation of our sentence-level approach to pragmatic tagging, that we revisit later.

\begin{figure}
    \centering
    \includegraphics[width=\linewidth]{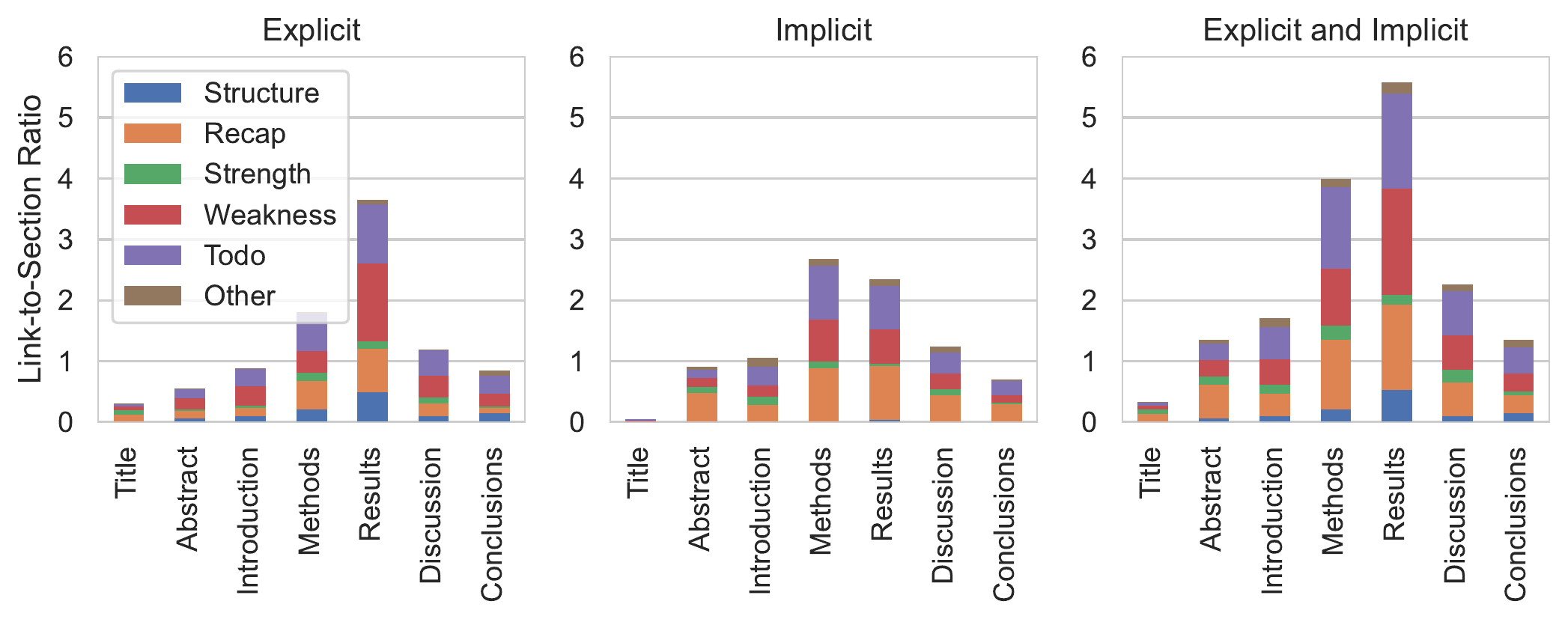}
    \caption{Pragmatic categories, links and paper sections. The number of links per general paper section section is normalized by the number of times the section appears in F1000RD.}
    \label{fig:joint_link_paper}
\end{figure}

\paragraph{What gets discussed} Just as the link anchors depend on peer review pragmatics, one could assume that link targets depend on the pragmatics of the papers under review -- not each section is equally likely to be addressed by the reviewers. Unlike free-form peer review reports, the pragmatics of research publications is explicitly signalled by paper structure, and for this experiment we have mapped the section titles of F1000RD sample to few groups common to academic publishing -- title, abstract, introduction, methods, results, discussion and conclusions. We then calculated the number of incoming links that each of those sections accumulates normalized by the number of times the section name appears in F1000RD. Figure \ref{fig:joint_link_paper} shows the results of our analysis along with the pragmatic category distribution on the peer review side. As we can observe, the most linked-to sections are Results and Methods -- yet, the distribution of linked-from statements from the peer reviews also differs depending on the paper section and the type of linking -- for example, while Abstracts are a frequent target of implicit \texttt{Recap}, Results are often explicitly referenced by \texttt{Weakness} sentences. While deeper analysis of those interactions lies beyond the scope of this work, we note the depth of analysis enabled by combining different intertextual signals.

\begin{figure}
    \centering
    \includegraphics[width=\linewidth]{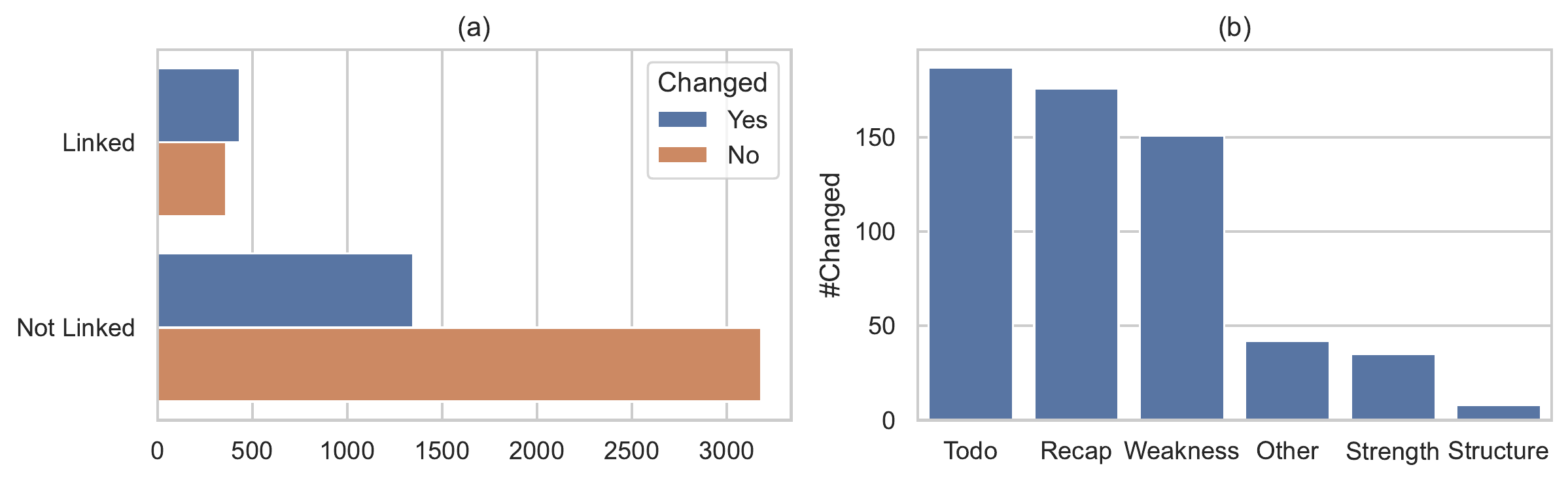}
    \caption{Left: Linking and node change in F1000RD data. Right: distribution of review sentence pragmatics for sentences linked to the paper paragraphs that were updated in a revision.}
    \label{fig:joint_change_910}
\end{figure}

\paragraph{What triggers change} Finally, the version alignment annotations allow us to study the effects of peer reviewing on manuscript revision. We first analyse the interaction between linking and changes: as Figure \ref{fig:joint_change_910} (left) shows, while most paper sentences are not linked from the peer review, the linked ones -- i.e. the ones discussed by the review -- are almost twice as likely to be changed in a subsequent revision (probability of change is $0.55$ and $0.30$ for paragraphs with and without incoming links, respectively). On the review side, we analyze what kinds of peer reviewer statements tend to trigger change (Fig. \ref{fig:joint_change_910}, right), and find that among the linked-to paragraphs of the paper, the most impact on the manuscript revision comes from the \texttt{Todo} sentences, followed by \texttt{Recap} and \texttt{Weakness} -- while other pragmatic categories are only responsible for few revisions. We note that our analysis only provides a coarse picture of reviewing and revision behaviour due to the differences in granularity and the intrinsic issues associated with sentence-level modeling of pragmatics; for example, a high number of changes triggered by the supposedly neutral Recap sentences points at potential limitations of our model. While our proposed framework allows a more fine-grained investigation, here we leave it for the future.

\section{Discussion}
\label{sec:discussion}

Having reviewed the instantiation of our proposed text-based collaboration model in the peer reviewing domain, we now take a step back and discuss the implications of our results specific to peer review, and in general. Our proposed model is centered around three core intertextual phenomena: pragmatic tagging is an instance of architextuality -- the relationship between a text and its abstract genre prototype; implicit linking is a reflection of metatextuality -- the relationship between a text and its commentary; finally, version alignment taps into the paratextuality -- by modeling the relationship between a text and its versions. Our proposed general view on intertextuality in NLP allowed us to systematize cross-document relationships along few basic axes -- explicitness, type and granularity -- surfacing connections between previously unrelated phenomena and highlighting the gaps that our subsequent study in the peer reviewing domain aims to fill. Modeling different phenomena within a joint framework makes it easy to study interactions between different types of intertextuality, and allows us to explore new research questions. To support our study, we propose a novel graph-based data model that is well suited for multi-document, relation-centered analysis on different granularity levels. The model generalizes over the common structural elements used to encode texts and can be easily extended to new domains and source types -- while the current study has focused on the data from F1000Research, our current implementation includes converters for Wikipedia, with the support for more text sources on the way. Our studies in peer reviewing discourse provide valuable, open, cross-domain datasets for the development of novel NLP applications for peer reviewing support. We deliver insights that can help shape future studies in modeling text-based collaboration, and we now briefly outline our main take-aways.

\paragraph{Data model} The main motivation behind our proposed graph-based data model is to facilitate modeling of intertextual relations on different granularity levels -- e.g. from sentence to paragraph. Yet, a general representation of document structure and other non-linguistic signal that surrounds texts has other potential uses, incl. language model fine-tuning and easy experimentation in cross-domain modeling; in addition, a graph-based data model allows preserving multimodal content -- like tables or figures -- by encapsulating it in the corresponding nodes, available for future processing. The further development of the Intertextual Graph is to address three key challenges. As support for more input formats will be added, the data model definition is likely to be refined. To allow massive LM pre-training and fine-tuning, computational overhead associated with additional processing needs to be addressed. Finally, new conceptual solutions are necessary to efficiently utilize the additional information encoded by Intertextual Graphs in modern Transformer-based NLP setups addressing particular end tasks.

\paragraph{Pragmatic tagging} Our approach to pragmatic tagging as a classic sentence labeling task has shown good results -- the annotations are reliable and the schema provides good coverage for the discourse analysis of peer reviews. While our label set is tailored towards peer reviews, alternative schemata can be developed or adapted from prior work to cover new text genres, e.g. scientific publications or news. Within the peer reviewing domain, our analysis suggests few additional pragmatic classes to be included in future versions of the labeling schema, most prominently, the performatives (\textit{``I thereby accept this paper''}) and verbose confidence assessments (\textit{``I'm not an expert in this area, but...''}). While in this work we resorted to sentence-level granularity for simplicity, a \emph{clause} provides a perhaps more natural unit of analysis for discourse tagging, e.g. as in \citet{ampere}. We leave exploring the effects of granularity on annotation speed and quality for the future, noting that the ITG data model readily supports sub-sentence granularities.

\paragraph{Linking} Our analysis of linking behavior in the peer reviewing domain has revealed that the two distinct linking mechanisms -- explicit and implicit linking -- differ substantially in terms of annotation and processing complexity. Explicit linking can be largely tackled by simple rule-based approaches -- facilitated by the ability of ITG to represent different structural elements of text. Yet, implicit linking requires more than that, as the annotation task has proven hard, reaching agreement levels similar to related studies \cite{radev-etal-2004-cst} and other challenging tasks like argumentation mining \cite{argmin3}. Our additional experiments involving re-test and expert annotations revealed that neither domain expertise nor subjectivity appear to affect the agreement -- yet the lack of such comparisons in related work prevents us from making general conclusions about the nature of the task. We note that even our domain-expert annotators were neither authors nor reviewers of the annotated publications. Soliciting implicit linking annotations from the reviewers and authors presents a viable, yet organizationally challenging alternative to producing labels in external annotation studies. From the annotation perspective, we deem it promising to explore the effects of the annotation interface and suggestion methods on the annotation quality and efficiency. From the task definition perspective, a promising alternative to our binary labeling schema is a decompositional approach where instead of producing a single binary label, annotators would answer a range of questions about the relationship between the anchor sentence and the target document, in the spirit of decompositional semantics \cite{decomp} applied to challenging sub-sentential phenomena. We leave this exploration to future work.

\paragraph{Version alignment} We have proposed a simple ITG alignment technique that does not require supervision, and -- as our results demonstrate -- provides good-quality paragraph-level alignments of F1000Research article revisions. Our proposed method is flexible and allows incorporating additional logical restrictions into the alignment process via ILP constraints. Yet it is important to note that the version alignment problem is far from solved: despite the high precision scores, only 70\% of the documents in our study were aligned perfectly. Moreover, the revision practices might vary substantially among research communities and publishing platforms. This might make the direct application of our proposed method problematic -- for example, as F1000Research does not put a size limitation on the publications, many revisions grow incrementally -- yet a page limit might potentially increase the number of modifications, as well as splitted and merged paragraphs, which are currently not supported by our aligner. Furthermore, while paragraph-level granularity has proven sufficient in our analysis, it might be insufficient for other applications. We deem it important to determine the parameters that affect revision practices across application scenarios and communities, and to collect diverse corpora of long-scope document revisions to support further investigation of the version alignment task.

\paragraph{Joint modeling} Our final study in joint modeling of peer reviewing discourse has demonstrated the advantages of an integrated approach to text-based collaboration within the proposed data model. While the results reported here are illustratory and much deeper analysis is possible, we note that some limitations of the proposed approaches only become evident when the tasks are considered jointly. Our analysis of the reviewers' linking behaviour revealed that an additional mechanism for modeling linking scope could be beneficial -- while only mentioned in a single sentence, a link might in fact connect the whole subsequent segment of a text to a location in another text. Whether linking scope should be modeled as part of pragmatic tagging and segmentation or as a separate information layer remains an open question. The optimal granularity level for the analysis of linking and revision behavior demands future investigation, as well.

\section{Conclusion}
\label{sec:conclusion}

Text-based collaboration is at the core of many processes that shape the way we interact with information in the digital age. Yet the lack of general models of collaborative text work prevents the systematic study of text-based collaboration in NLP. To address this gap, in this work we proposed a model of text-based collaboration inspired by related work in intertextuality theory, and introduced three general tasks that cover one full cycle of document revision: pragmatic tagging, linking, and version alignment. To support our study, we developed the Intertextual Graph -- a generic data model that takes place of the ad-hoc plain text-based representation and is well suited for describing long documents and cross-document relations. We investigated the application of the proposed model to peer reviewing discourse and created F1000RD -- the first clearly licensed, multi-domain NLP corpus in open post-publication peer review. Our annotation studies revealed the strengths and weaknesses of our proposed approaches to pragmatic tagging, linking and version alignment, and allowed us to determine promising directions for future research. While our proposed framework for analysis of text-based collaboration in NLP is joint -- covering different aspects of this collaboration -- a further necessary prerequisite for generalization is its unification across applications and domains. Thus, along with refining the task definitions and developing NLP models for performing the annotation tasks automatically, we deem it crucial to expand our proposed framework to new scenarios -- by creating new data and incorporating existing data sources from other key domains of text-based collaboration like Wikipedia, news, online discussion platforms, and others.

\starttwocolumn
\bibliography{compling_style}

\end{document}